%% file: MixLRCo.tex
\documentclass[10pt,twocolumn,letterpaper]{article}

\usepackage[pagenumbers]{cvpr} 

\usepackage{graphicx}
\usepackage{amsmath}
\usepackage{amssymb}
\usepackage{booktabs}

\usepackage{multirow}
\usepackage{graphicx}
\usepackage{booktabs}
\usepackage{setspace}
\usepackage{bm}
\usepackage{bbm}
\usepackage{bbding}
\usepackage{pifont}
\newcommand{\ltwonorm}{\ell_{2}}

\usepackage[pagebackref,breaklinks,colorlinks]{hyperref}

\usepackage[capitalize]{cleveref}
\crefname{section}{Sec.}{Secs.}
\Crefname{section}{Section}{Sections}
\Crefname{table}{Table}{Tables}
\crefname{table}{Tab.}{Tabs.}

\begin{document}

\title{Low-confidence Samples Matter for Domain Adaptation}

\author{Yixin~Zhang\qquad~Junjie~Li\qquad{Jiafan Zhuang}\qquad~Zilei~Wang\thanks{Corresponding author}\\
Department of Automation, University of Science and Technology of China\\
{\tt\small \{zhyx12, hnljj, jfzhuang\}@mail.ustc.edu.cn} \qquad {\tt\small zlwang@ustc.edu.cn}
}


\maketitle

\begin{abstract}
Domain adaptation (DA) aims to transfer knowledge from a label-rich source domain to a related but label-scarce target domain. The conventional DA strategy is to align the feature distributions of the two domains. Recently, increasing researches have focused on self-training or other semi-supervised algorithms to explore the data structure of the target domain. However, the bulk of them depend largely on confident samples in order to build reliable pseudo labels, prototypes or cluster centers. Representing the target data structure in such a way would overlook the huge low-confidence samples, resulting in sub-optimal transferability that is biased towards the samples similar to the source domain. To overcome this issue, we propose a novel contrastive learning method by processing low-confidence samples, which encourages the model to make use of the target data structure through the instance discrimination process. To be specific, we create positive and negative pairs only using low-confidence samples, and then re-represent the original features with the classifier weights rather than directly utilizing them, which can better encode the task-specific semantic information. Furthermore, we combine cross-domain mixup to augment the proposed contrastive loss. Consequently, the domain gap can be well bridged through contrastive learning of intermediate representations across domains. We evaluate the proposed method in both unsupervised and semi-supervised DA settings, and extensive experimental results on benchmarks reveal that our method is effective and achieves state-of-the-art performance. The code can be found in \url{https://github.com/zhyx12/MixLRCo}.

\end{abstract}

\input{sec_introduction}

\input{sec_related_work}

\input{sec_preliminary}

\input{sec_method}

\input{sec_experiments}

\section{Conclusion}

\vspace{-2mm}
In this paper, we propose a novel contrastive learning framework for unsupervised domain adaptation (UDA) and semi-supervised domain adaptation (SSDA) by exploring the low-confidence samples in the target domain and making use of classifier weights to re-represent the features.
Our proposed method can effectively alleviate the semantic conflict problem of the original contrastive loss.
Furthermore, we propose to integrate the cross-domain mixup into contrastive learning, which can help to reduce the domain gap and further boost the performance.
Extensive experiments show the effectiveness of the proposed method, which can achieve state-of-the-art UDA and SSDA performance.

{\small
\bibliographystyle{ieee_fullname}
\bibliography{MixLRCo}
}

\end{document}

%% file: sec_introduction.tex
\section{Introduction}

\begin{figure}[!t]
	\centering
	\includegraphics[width=0.98\columnwidth]{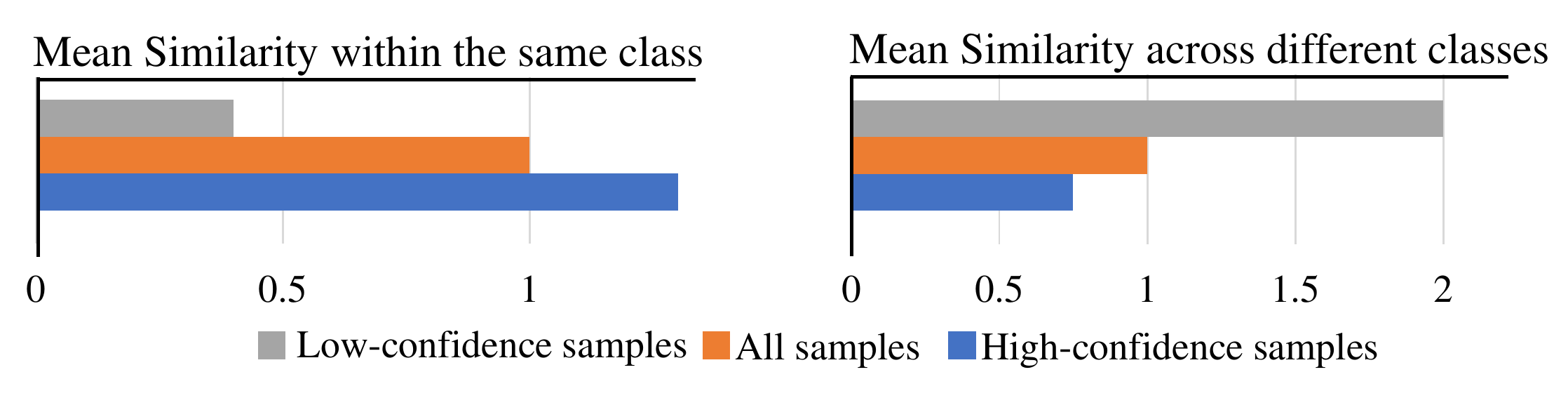}
	\caption{\small Mean similarity within the same class (left) and across classes (right). In each figure, mean similarities of high-confidence, low-confidence and all samples are considered, and these values are scaled to make the one of all sample equal 1 for better comparison.}
	\label{fig:motivation}
	\vspace{-3mm}
\end{figure}

Deep neural networks (DNNs) have shown considerable effectiveness in a variety of machine learning challenges~\cite{krizhevsky2012alexnet,chen2017deeplab,ren2015faster}. However, the impressive performance gain heavily relies on the access to massive well-labeled training data. Additionally, manually annotating sufficient training data is often time and expense prohibitive in reality. Besides, another disadvantage of traditional deep learning is its inability to generalize to new datasets due to the domain shift problem \cite{ben2007analysis, Ben2010A}. Domain adaptation (DA) address this issue by utilizing the knowledge of a label-rich domain (\textit{i.e.}, source domain) to assist the learning in a related but label-scarce domain (\textit{i.e.}, target domain). We investigate two scenarios in this paper: unsupervised DA (UDA) and semi-supervised DA (SSDA). The distinction is that the target domain has no labeled data (UDA) or only a small quantity of labeled data (SSDA).

The most popular way to deal with the issue of domain shifting is to learn domain-invariant representations. We can classify these DA approaches as either discrepancy metric-based~\cite{tzeng2014deep, long2017deep, yan2017mind} or adversarial-based~\cite{ganin2015dann, tzeng2017adversarialadda, liu2019transferabletat, saito2018maximummcd, cui2020gvb}. Recently, there have been more methods exploring the inherent structures of unlabeled target domains, such as self-training through pseudo labels~\cite{french2018selfensemble, zou2018CBST, kim2019selftraining, liu2021cycle_selftraining} and aligning prototypes across domains~\cite{pan2019TPN, chen2019PFAN,CAN2019,zhang2021prototypical}.
To create reliable pseudo labels or prototypes in the target domain, the majority of them simply pick trustworthy samples according to some criteria, such as probability~\cite{pan2019TPN, chen2019PFAN} and sample ratio~\cite{zou2018CBST}. Consequently, the low-confidence samples are completely ignored in these methods.
There are some approaches taking into account less reliable samples, such as self-penalization loss~\cite{na2021fixbi} and entropy minimization~\cite{zhang2019domainsymnet, gu2020sphericalDA, xu2019largernorm, roy2019feature-whitening, mei2020iast}. However, these methods only consider the probability or entropy of each individual sample and disregard the connections of these samples.
%



Most methods does not process these low-confidence samples due to no obvious supervision signal such as pseudo-label. In light of the recent success of instance discrimination contrastive loss ~\cite{he2020momentum,chen2020simple} in unsupervised representation learning, one natural idea is to introduce it to the unlabeled target domain since it does not need semantic labels. It should be noted that most existing contrastive learning methods in domain adaptation are based on classes, where a sample should be close to those of the same category, and away from those of other categories. Thus, only high-confidence samples are used by resorting to pseudo labels. We use instance discrimination contrastive loss which is based on individual instances. However, as stated by previous works~\cite{yue2021pcs,kim2020cross_domain_ssl}, the instance discrimination has a basic flaw for domain adaptation: the semantic structure of the data is not encoded by the learned representations. This is because two instances, regardless of their semantics, are considered as negative pairs if they originate from different samples, and their similarity will be reduced.
To solve this problem, we delve into the feature similarities of samples within the same class and across classes. Fig.~\ref{fig:motivation} depicts our findings. Specifically, for the low-confidence samples, the mean similarity within a class is smaller and that across classes is larger compared with the high-confidence samples. This phenomenon about the low-confidence samples is opposite to the ideal case (\ie, similarity within the same class is larger, and that across classes is smaller). But it may be appropriate for contrastive learning: given randomly sampled negative pairs, the contrastive loss will pay more attention to the samples across classes due to larger similarities, and will push them away from each other. Thus we propose to only use low-confidence samples for contrastive loss, which would alleviate the semantic conflict in contrastive learning. To further incorporate the task-specific semantic information, we propose to re-represent the original feature with the classifier weights that work as a new set of coordinates. 
In this paper, we denote the proposed \textbf{L}ow-confidence sample based feature \textbf{R}e-representation \textbf{Co}ntrastive Loss by the abbreviation LRCo.

While LRCo primarily concerns the data structure of the target domain, we further propose to bridge the domain gap by combining it with cross-domain mixup~\cite{zhang2018mixup}, namely, MixLRCo. Here the model is encouraged to behave linearly across the source domain and target domain samples.
This way is inspired by recent advances in self-supervised learning~\cite{shen2020unmix, lee2021imix,kalantidis2020hardmix} that use mixup to generate positive or negative samples. In this work, we only use low-confidence target samples and random source samples for cross-domain mixup. What is more, the low-confidence target samples are given a higher weight than the source samples, which can keep the attention of contrastive learning to the data structure of the target domain.

Our contributions are summarized as follows:
\vspace{-2mm}
\begin{itemize}
    \setlength\itemsep{-0.2em}
	\item We propose a novel domain adaptation framework, LRCo, that explores the inherent structures of the target domain by performing contrastive learning on low-confidence samples. Additionally, a simple feature re-representation method is also proposed to encode the task-specific information.
	\item We extend LRCo by combining with the cross-domain mixup, which can reduce the domain discrepancy by producing intermediate samples across domains and contrastively learning their representations.
	\item We conduct extensive experiments on multiple UDA and SSDA benchmarks, and the results reveal that our proposed method achieves the state-of-the-art performance owing to better exploiting the low-confidence samples.
\end{itemize}

%% file: sec_related_work.tex
\section{Related Work}
\subsection{Unsupervised Domain Adaptation}
Unsupervised Domain Adaptation (UDA)~\cite{ben2007analysis, Ben2010A} is a technique for generalizing a model learnt from a labeled source domain to an unlabeled target domain. The mainstream approaches are to learn domain-invariant representations, and they can be classified into two coarse types. The first one specifically decreases the domain discrepancy represented by some distribution discrepancy metrics~\cite{tzeng2014deep, long2017deep, yan2017mind, sun2016return, sun2016deep, peng2019moment}. Another common line of research is adversarial training~\cite{ganin2015dann, tzeng2017adversarialadda, liu2019transferabletat, wang2019transferableattention, saito2018maximummcd, cui2020gvb, Li_2021_ICCV_SCDA}. These methods make use of all samples in both the source and 
target domains, which usually serve as a baseline for other advanced methods.
To increase discriminability, more recent DA methods attempt to investigate the data structure in unlabeled target domain. Self-training as a typical approach generates target domain pseudo labels~\cite{saito2017asymmetric_tri, french2018selfensemble, zou2018CBST, li2019bidirectional, zou2019crst, mei2020iast,pan2020unsupervised_intra_da, kim2019selftraining, zhang2021rpa, zhang2018iCAN, kumar2020gradualDA, na2021fixbi,liu2021cycle_selftraining}. Another category is to construct the protoptypes~\cite{pan2019TPN, chen2019PFAN, xie2018mstn, zheng2020coarse_to_fine, xu2020gpa, zhang2021prototypical} or cluster centers~\cite{CAN2019, deng2019clusterwithteacher, tang2020srdc} across domains and then perform class-wise alignment. Most of these approaches use a fixed probability threshold~\cite{saito2017asymmetric_tri,french2018selfensemble, pan2019TPN, li2019bidirectional, xu2020gpa}, a dynamic probability threshold~\cite{zhang2018iCAN, na2021fixbi}, a fixed sample ratio~\cite{kumar2020gradualDA, pan2020unsupervised_intra_da, zhang2021rpa}, a dynamic sample ratio~\cite{saito2017asymmetric_tri, zou2018CBST, zou2019crst}, or a threshold of other metrics ~\cite{gu2020sphericalDA, chen2019PFAN, kim2019selftraining,liu2021cycle_selftraining} to choose trustworthy samples (\ie, high-confidence samples) and reject other low-confidence samples.
To mitigate the harmful effect of noisy labels, it is reasonable to only utilize the reliable samples. However, less trustworthy samples are also important since they may reveal the structure of data space.
Some techniques are proposed to maintain a global prototype bank~\cite{deng2019clusterwithteacher, CAN2019, zheng2020coarse_to_fine}. The entropy weighted methods~\cite{long2018CDAN,wang2019transferableattention} give greater weights to the samples with low entropy (\ie, high-confidence) and lower weights to the samples with high entropy (
\ie, low-confidence). In both circumstances, the influence of incorrect predictions is assumed to be automatically reduced by the majority of correct predictions. But the explicit supervisions for low-confidence samples are still lacking.

Some approaches try to add more constraints to low-confidence samples. FixBi~\cite{na2021fixbi} proposes a self-penalization loss that enforces the maximum probability of the low-confidence samples to be zero. Low-confidence samples are also taken into account by entropy minimization~\cite{zhang2019domainsymnet, vu2019advent, gu2020sphericalDA, xu2019largernorm, roy2019feature-whitening, mei2020iast} and virtual adversarial training~\cite{cicek2019rca, shu2018dirt}. These constraints are built based on the information (\eg, probability and entropy) of individual samples, while neglecting their relationships. Some methods~\cite{ jin2020MCC,li2021BCDM,Li_2021_ICCV_SCDA} propose to use dark knowledge~\cite{hinton2015distilling} (\ie, knowledge on incorrect DNN predictions), but their primary objective is to suppress incorrect predictions through inconsistency discovery. Different from these methods, we propose to learn the representation of low-confidence samples more effectively via an instance discrimination procedure.

\vspace{-1mm}

\subsection{Semi-Supervised Domain Adaptation}

Semi-Supervised Domain Adaptation (SSDA) aims to reduce the discrepancy between the source and target distribution in the presence of limited labeled target samples. ~\cite{saito2019mme} first proposes to align the distributions using adversarial training on entropy.
Based on MME, ~\cite{Kim2020AttractPA,2020Bidirectional} further introduce virtual adversarial training~\cite{miyato2018VAT} to learn more robust features.
\cite{li2020online} presents a meta-learning framework for SSDA. 
\cite{yang2021decota} breaks down the SSDA problem into two sub-problems, namely, semi-supervised problem and UDA problem, and then proposes to learn consistent predictions using co-training.
CDAC~\cite{li2021cdac} proposes an adversarial adaptive clustering loss to group features of unlabeled target data into clusters and perform cluster-wise feature alignment across domains.
CLDA~\cite{singh2021clda} employs class-wise contrastive learning to reduce the inter-domain gap and instance level contrastive alignment to minimize the intra-domain discrepancy.
ECACL~\cite{Li_2021_ICCV_ecacl} propose a holistic framework that incorporates multiple complementary domain alignment techniques. Among these methods, ~\cite{li2021cdac, mishra2021surprisinglyss, Li_2021_ICCV_ecacl} adopt FixMatch~\cite{fixmatch} or its variants. In this work, we also use FixMatch to build a strong baseline, and add a regularization term~\cite{zou2019crst} to get more reliable pseudo labels.

\subsection{Contrastive Representation Learning}
Contrastive Learning has shown remarkable advantages in self-supervised learning~\cite{he2020momentum,chen2020simple,dwibedi2021little,misra2020self,hjelm2019learning,oord2018representation}. The contrastive loss measures the similarity of representation pairs and attempts to distinguish between positive and negative pairs.
%
%
MoCo~\cite{he2020momentum} maintains a queue of previously processed embeddings as negative memory bank.
SimCLR~\cite{chen2020simple} shows that large batch size and strong data augmentations has a comparable performance to the memory-based approaches.
Here we adopt a similar architecture to MoCo~\cite{he2020momentum} to perform contrastive learning.
%
%

%
In domain adaptation, most approaches~\cite{CAN2019, yue2021pcs} use contrastive loss on the basis of class-wise prototypes with a sample selection strategy.
Only a few methods~\cite{kim2020cross_domain_ssl, sahoo2021contrastmix, singh2021clda} use the instance discrimination based contrastive loss.
\cite{kim2020cross_domain_ssl} employs contrastive learning in a pre-train step before proceeding to a domain alignment stage.
\cite{singh2021clda} suggests that the classifier can be used as a contrastive projection head~\cite{chen2020simple}.
\cite{sahoo2021contrastmix} conducts temporal contrastive self-supervised learning over the graph representations.
In contrast to previous efforts, we particularly concentrate on contrastive learning on low-confidence samples and at the same time propose a feature re-representation approach for encoding the task-specific semantic information.

\subsection{Mixup Training Strategy}
Mixup~\cite{zhang2018mixup} and its variants~\cite{berthelot2019mixmatch, verma2019manifoldmix, yun2019cutmix} provide effective data augmentation strategies when paired with a cross-entropy loss for supervised and semi-supervised learning. 
In domain adaptation, there are several methods~\cite{xu2020domainmixup, yan2020improvemixup, mao2019virtualmixup, wu2020dualmixup, na2021fixbi, yang2021decota} applying mixup. \cite{xu2020domainmixup, wu2020dualmixup} combines the domain-level mixup with domain adversarial training to learn a more continuous latent space across domains. \cite{yan2020improvemixup, na2021fixbi, yang2021decota} use the category-level mixup which requires the target domain pseudo labels. \cite{mao2019virtualmixup} proposes virtual mixup training by combining virtual adversarial training and mixup. 
In self-supervised learning, recent works have leveraged the idea of image space mixtures~\cite{shen2020unmix, lee2021imix} and embedding space mixtures~\cite{kalantidis2020hardmix, zhu2021featuretransformation} to generate more valuable positive or negative samples. Our MixLRCo is inspired by these methods but implemented in a non-trivial way. Additionally, our approach is different from ~\cite{sahoo2021contrastmix} that adopts the background mixing for video domain adaptation and treat each domain equally.


%% file: sec_preliminary.tex
\section{Preliminary}

In this work, we focus on the unsupervised and semi-supervised domain adaptation problem in image classification.
Formally, for both tasks, we are given a source domain $\mathcal{D}_s = \{(\bm{x}^i_s,{y}_s^i)\}_{i=1}^{n_s}$ with $n_s$ labeled samples and an unlabeled target domain ${{\cal D}_{tu}} = \{ {\bm{x}}^j_{tu}\} _{j = 1}^{{n_{tu}}}$ with $n_{tu}$ unlabeled samples, except in the semi-supervised setting, a small amount of extra labeled samples in the target domain are given
${{\cal D}_{tl}} = \{ {\bm{x}}^k_{tl}, {y}^k_{tl}\}_{k = 1}^{{n_{tl}}}$. The data in source and target domains are drawn from the joint distributions $P(\bm{x}_{s}, y_{s})$ and $Q(\bm{x}_{t}, y_{t})$ with $P \ne Q$, respectively.

%
Inspired by~\cite{chen19closerfewshot, gu2020sphericalDA, saito2019mme, yue2021pcs}, we build a cosine similarity based network. It consists of a feature extractor $F$ and a classifier $C$.
The output probability can be obtained by ${\small \mathbf{p(x)}} = \sigma(\small{\frac{\mathbf{W}_C^{\mathrm{T}}\ltwonorm(F(\mathbf{x}))}{T_{ce}}})$,
%
where $\sigma{}$ indicates a softmax function, $\mathbf{W}_C$ is classifier weights, $\ltwonorm(\mathbf{x}) = \frac{\mathbf{x}}{\|\mathbf{x}\|}$ is the $\ell_2$- normalization function, and $T_{ce}$ is a temperature parameter.


Given this architecture, we build our method on the existing well-performed baseline. Generally speaking, the training objective can be summarized as 

\vspace{-4mm}

\begin{equation}
    \begin{aligned}
        \ell_{ce} &=- \sum_{i=1}^{K} y_{i} \log p(y=i|\mathbf{x}), \\
        \mathcal{L}_{baseline} &= \mathbb{E}_{\mathbf{x} \in \mathcal{D}_{l}} \ell_{ce} + \lambda_{align} \mathbb{E}_{\mathbf{x}\in {\mathcal{D}_{l} \cup \mathcal{D}_{tu}} }  \ell_{align},
    \end{aligned}
	\label{eq:baseline}
\end{equation}

\noindent where $\mathcal{D}_{l}$ means labeled domain, and represents $\mathcal{D}_{s}$ in UDA and $\mathcal{D}_{s} \cup \mathcal{D}_{tl}$ in SSDA. $K$ is the number of classes.
$\ell_{ce}$ is the standard cross-entropy loss, and $\ell_{align}$ is some specific loss to align representations across domains. Since forms of different baselines differ, we don't give specific definition for $\ell_{align}$. $\lambda_{align}$ is the trade-off parameter.

%% file: sec_method.tex
\section{Method}



\subsection{A Stronger Baseline}
\label{sec:strong_baseline}

Inspired by previous works~\cite{zhang2021sslanduda, zhang2021prototypical, li2021cdac} in domain adaptation, we add FixMatch~\cite{fixmatch} to the existing method to construct a stronger baseline. Specifically, let $\mathcal{T}(\mathbf{x})$ and $\mathcal{T^\prime}(\mathbf{x})$ denote the weakly and strongly augmented views for $\mathbf{x} \in \mathcal{D}_{tu}$, respectively. To generate more stable pseudo labels, we adopt a teacher-student framework.
The teacher model (\ie, $\tilde{F}$ and $\tilde{C}$) is continuously updated by exponential moving average (EMA)~\cite{tarvainen2017mean} of the student model (\ie, $F$ and $C$). Here we use weight decay rate $0.99$ for all experiments.


%

To obtain a pseudo label, we first compute probability of a weak augmented view: $\mathbf{\tilde{p}(\mathcal{T}(\mathbf{x}))}$. Then we use $\hat{p} = \mathtt{argmax}(\mathbf{\tilde{p}})$ as the pseudo label. The loss on the strongly augmented view can be presented as 

\vspace{-3mm}
\begin{equation}
	\ell_{fm} = -\mathbbm{1}(\max(\mathbf{\tilde{p}})>\tau)  \log \mathbf{p}(y=\hat{p}|\mathcal{T'}(\mathbf{x})),
	\label{eq:fixmatch}
\end{equation}
\noindent where $\tau$ is the probability threshold. We can split the data in target domain $D_{tu}$ into $D_{tu}^{h}$ and $D_{tu}^{l}$ at each training step. The superscript $h$ and $l$ are used to denote the high-confidence and low-confidence samples, respectively.

Although FixMatch improves the prediction consistency and provide more reliable pseudo labels, the wrong predictions are still severe due to domain discrepancy~\cite{liu2021cycle_selftraining}. Here we follow~\cite{mei2020iast, zhang2021prototypical} to use the KLD regularization term in CRST~\cite{zou2019crst}. It encourages the high-confidence output to be evenly distributed to all classes so that the prediction results do not overfit the pseudo labels.

\vspace{-3mm}

\begin{equation}
	\ell_{kld} = - \mathbbm{1}(\max(\mathbf{\tilde{p}})>\tau) \sum_{j=1}^{K} \frac{1}{K} \log \mathbf{p}(y=j|\mathcal{T'}(\mathbf{x}_{tu})).
	\label{eq:kld}
\end{equation}

\noindent Here $\ell_{kld}$ is only applied to the high-confidence samples. Thus the overall loss function of our strong baseline can be presented as follows. Here we set $\lambda_{kld} = 0.1$ for all experiments.

\vspace{-3mm}

\begin{equation}
	\mathcal{L}_{strong} = \mathcal{L}_{baseline} + \mathbb{E}_{\mathbf{x} \in \mathcal{D}_{tu}^h} (\ell_{fm} + \lambda_{kld} \ell_{kld}).
	\label{eq:strong_baseline}
\end{equation}

\begin{figure}[t]
	\centering
	\includegraphics[width=1.0\columnwidth]{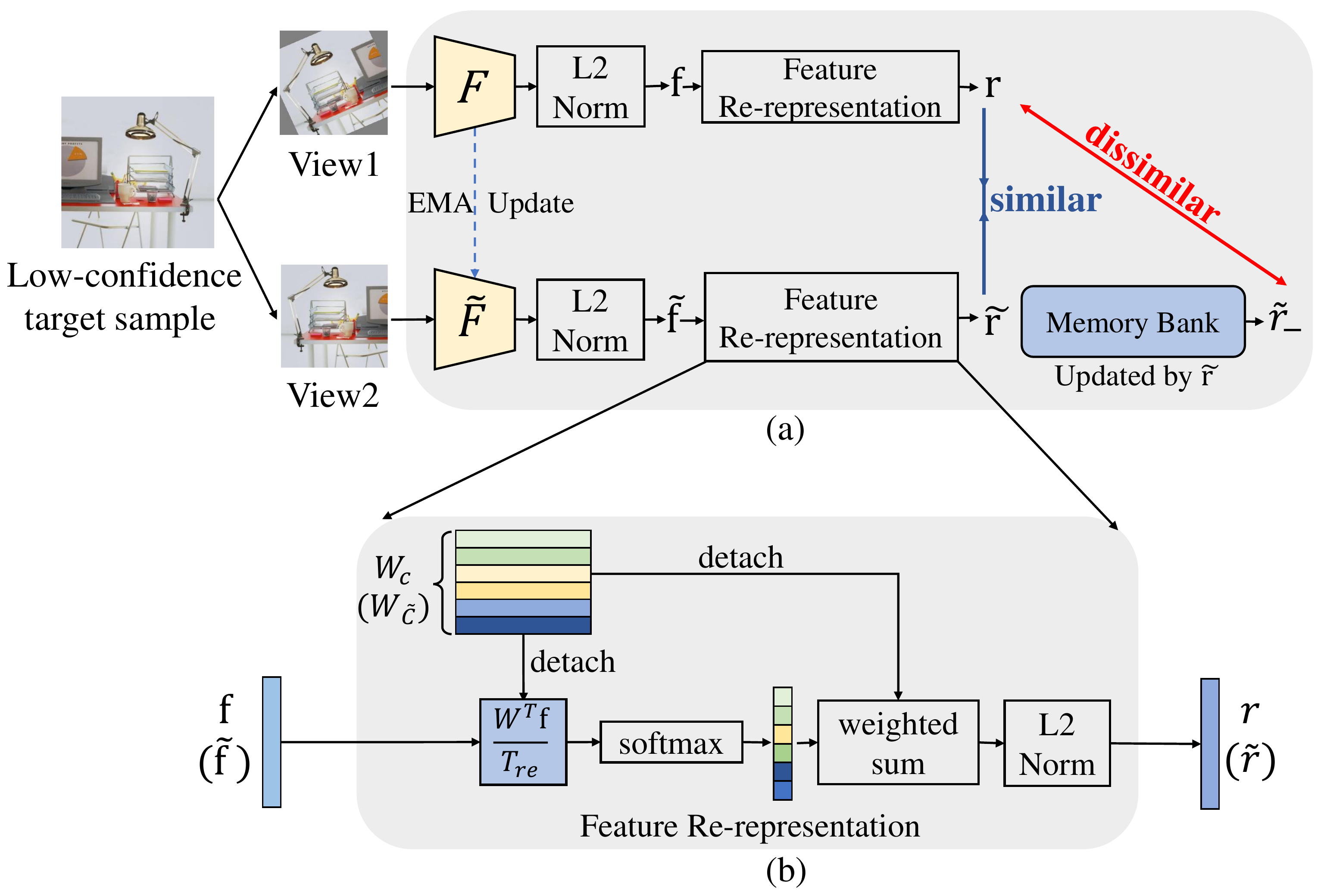}
	\caption{\small (a) The framework of our proposed LRCo. (b) The process of feature re-representation.}
	\label{fig:LRCo}
\end{figure}

\subsection{Contrastive Loss for Low-confidence Samples}
\label{sec:lrco}
Before introducing our new contrastive loss, we first review the contrastive loss used in self-supervised learning. Given an image $\mathbf{x}_{i} \in \mathcal{D}_{tu}$, we can obtain two differently augmented views $\mathbf{x}_{i}^{1}, \mathbf{x}_{i}^{2}$ as the query image and key image. Then the $\ell_2$-normalized features can be produced by $\mathbf{f}_{i} = \ltwonorm(F(\mathbf{x}_{i}^{1})), \mathbf{\tilde{f}}_{i} = \ltwonorm(\tilde{F}(\mathbf{x}_{i}^{2}))$. The naive contrastive loss without other designs (\eg, projection head) can be presented as follows:

\begin{equation}
	\begin{aligned}
		h(\mathbf{f}_i, \mathbf{\tilde{f}}_i) &= \exp (\mathbf{f}_i^T \mathbf{\tilde{f}}_i / T_{co}), \\
		\ell_{co} = - \log& \frac{h(\mathbf{f}_i, \mathbf{\tilde{f}}_i)}{ h(\mathbf{f}_i, \mathbf{\tilde{f}}_i) + \sum\limits_{\mathbf{\tilde{f}}_{-} \in M} h(\mathbf{f}_i,  \mathbf{\tilde{f}}_{-})},
	\end{aligned}
	\label{eq:orig_contrastive_loss}
\end{equation}

\noindent where $T_{co}$ is the temperature hyper-parameter for contrastive loss, and we use $h$ to denote the exponential of scaled cosine similarity. $M$ is the memory bank~\cite{he2020momentum} which stores the processed features by the teacher feature extractor. Intuitively, this loss is the log form of a ($|M| +1$)-way softmax-based classifier that tries to classify $\mathbf{f}_i$ as $\mathbf{\tilde{f}}_i$.

A drawback of the contrastive loss is that a pair of samples belonging to the same class could be treated as negatives. More specifically, for a balanced dataset, there are average $\frac{|M|}{K}$ samples in $M$ that belong to the same class as the query image but regarded as negative samples.
This can not be changed as long as $M$ is randomly constructed, even if only high-confidence or only low-confidence samples are considered\footnote{Here we assume that different classes are equally learned, and then the number of high (low)-confidence samples in each class are roughly equal at each training step.}.
However, as mentioned before, after inspecting the similarities of samples within same class and across classes, we argue that the low-confidence samples are more suitable for applying contrastive loss.


%

Instance discrimination contrastive loss works well in unsupervised representation learning, we tried directly applying it in feature space, but got limited improvement, which motivates us to encode task-specific semantic information.
%
%
Here we propose a new strategy by re-representing the original feature with the classifier weights, where an attention mechanism is used.

\vspace{-5mm}

\begin{equation}
	\begin{aligned}
        \mathtt{Re}(\mathbf{f}_{i}) = \sigma( \frac{\mathbf{W}_C^{\mathrm{T}}\mathbf{f}_{i}}{T_{re}})  \mathbf{W}_{C}, \quad
		\mathbf{r}_{i} = \ltwonorm(\mathtt{Re}(\mathbf{f}_{i})), \\
		\mathtt{Re}(\mathbf{\tilde{f}}_{i}) = \sigma( \frac{\mathbf{W}_{\tilde{C}}^{\mathrm{T}}\mathbf{\tilde{f}}_{i}}{T_{re}}) \mathbf{W}_{\tilde{C}}, \quad
		\mathbf{\tilde{r}}_{i} = \ltwonorm(\mathtt{Re}(\mathbf{\tilde{f}}_{i})),
	\end{aligned}
	\label{eq:re-representation}
\end{equation}
\noindent where $\mathbf{W}_{C}$ and $\mathbf{W}_{\tilde{C}}$ are the weights of student classifier and teacher classifier. $T_{re}$ is a temperature to scale the similarity, and we directly set $T_{re}=T_{ce}$ without over-tuning this hyper-parameter. It should be noted that the conclusion obtained from Figure~\ref{fig:motivation} still holds after feature re-representation. Here the classifier weights are regarded as the class prototypes, and we use them as a set of new coordinate to re-represent the original features. Thus the classifier weights are detached and not updated by the contrastive loss:


\vspace{-5mm}

\begin{equation}
	\begin{aligned}
		\ell_{lrco} = - \log& \frac{h(\mathbf{r}_i, \mathbf{\tilde{r}}_i)}{ h(\mathbf{r}_i, \mathbf{\tilde{r}}_i) + \sum\limits_{\mathbf{\tilde{r}}_{-} \in M} h(\mathbf{r}_i,  \mathbf{\tilde{r}}_{-})},
	\end{aligned}
	\label{eq:lrco}
\end{equation}
\noindent where $M$ stores the re-represented features.
Finally, the framework is shown in Fig.~\ref{fig:LRCo}, and the overall loss is presented as 
\begin{equation}
	\begin{aligned}
		\mathcal{L}_{lrco} = \mathcal{L}_{strong} + \lambda_{co} \mathbb{E}_{\mathbf{x} \in \mathcal{D}_{tu}^l}  \ell_{lrco}.
	\end{aligned}
	\label{eq:lrco_all}
\end{equation}

Our feature re-representation is similar to PRONOUN~\cite{hu2021pronoun} which adopts prototype-based normalized output. The differences are two-folds: \textbf{i}. motivation: they aim to enhance the conditioning strength when constructing input for domain adversarial training, we aim to encode semantic information and alleviate semantic conflict. \textbf{ii}. They maintain global prototypes, and we directly adopt classifier weight.

\subsection{Cross-Domain Mixup Contrastive Learning}
\label{sec:mixlrco}

\begin{figure}[!t]
	\centering
	\includegraphics[width=1.0\columnwidth]{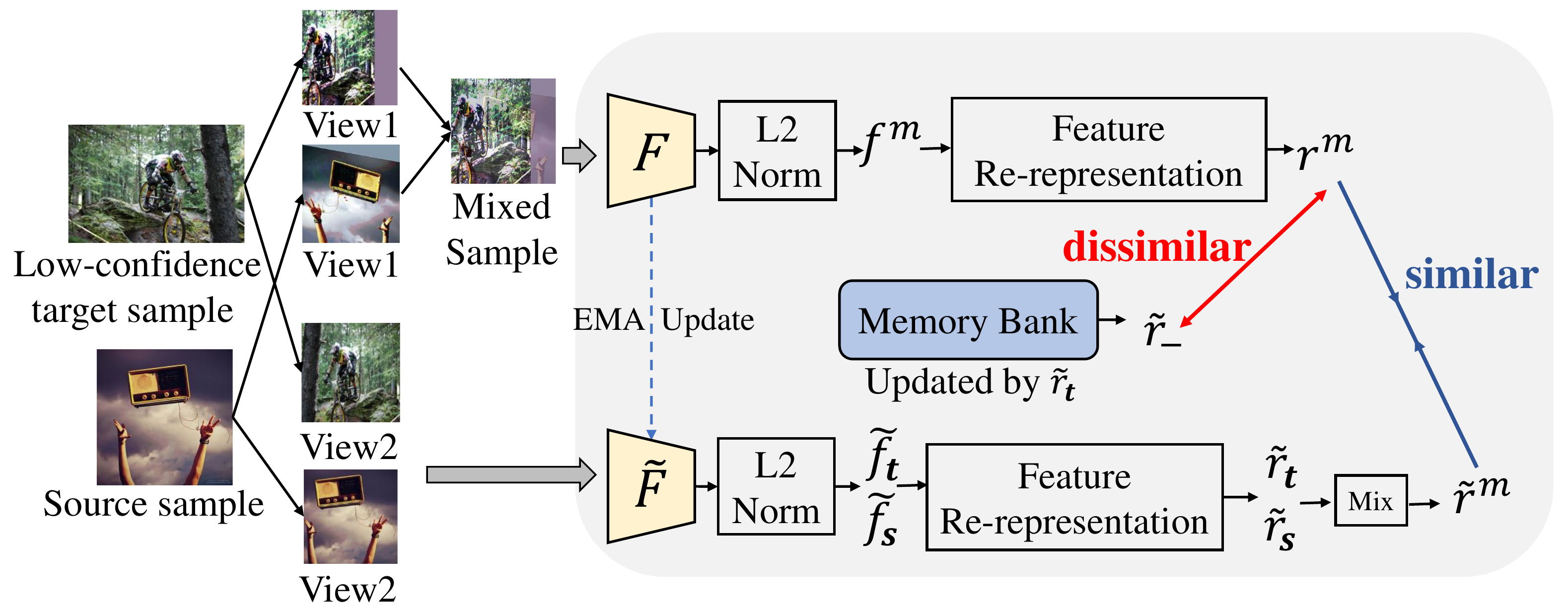}
	\caption{\small The framework of our proposed MixLRCo, which introduces the cross domain mixup into our LRCo.}
	\label{fig:MixLRCo}
	\vspace{-4mm}
\end{figure}

For the low-confidence samples in the target domain, one of the reasons why they are difficult to be correctly classified is the low similarity with the source domain samples. Meanwhile, the aforementioned contrastive learning process only considers the structure of feature space in target domain and ignores the domain differences. Inspired by recent work~\cite{kalantidis2020hardmix,lee2021imix,shen2020unmix, zhu2021featuretransformation} which combine mixup~\cite{zhang2018mixup} and contrastive learning, we propose cross-domain mixup contrastive learning to further boost the domain-shared feature learning as shown in Fig.~\ref{fig:MixLRCo}. Specifically, given two differently augmented views $\mathbf{x}_{i}^{1}, \mathbf{x}_{i}^{2}$ of the same image $\mathbf{x}_{i} \in \mathcal{D}_{tu}^l$, we randomly sample an image $\mathbf{x}_{s}$ from source domain, and also obtain two different views $\mathbf{x}_{s}^{1}$ and $\mathbf{x}_{s}^2$. Then we mixup the source samples and low-confidence target samples as query image. In order to make sure that the low-confidence target samples keep dominant, we adopt a similar strategy as in MixMatch~\cite{berthelot2019mixmatch} where a max operation is used upon the $\mathtt{Beta}$ distribution. Then the mixup process can be presented as 

\vspace{-3mm}

\begin{equation}
\begin{aligned}
	\lambda &\sim \mathtt{Beta}(\alpha, \alpha), \\
	\lambda^\prime &= \max(\lambda, 1 - \lambda),  \\
	\mathbf{x}^m_{i} &= \lambda^\prime \mathbf{x}_i^1 + (1 - \lambda^\prime)\mathbf{x}_s^{1}, \\
\end{aligned}
\label{eq:img_mixup}
\end{equation}
\noindent where $\alpha$ is the parameter controlling the shape of $\mathtt{Beta}$ distribution.

\input{./tables/uda_officehome}

Then the mixed image $\mathbf{x}_{i}^{m}$ is sent to the student feature extractor, and $\mathbf{x}_{i}^{2}$ and $\mathbf{x}_{s}^{2}$ are sent to the teacher feature extractor. To obtain the the query feature $\mathbf{r}_i^m$ and the key feature $\mathbf{\tilde{r}}^{m}$, we conduct the following process 

\vspace{-3mm}

\begin{equation}
	\begin{aligned}
		\mathbf{f}_{i}^{m} = \ltwonorm(F(\mathbf{x}_{i}^{m})), & \quad
		\mathbf{r}_{i}^{m} = \ltwonorm(\mathtt{Re}(\mathbf{f}_{i})), \\
		\mathbf{\tilde{f}}_{i} = \ltwonorm(\tilde{F}(\mathbf{x}_{i}^{2})),& \quad
		\mathbf{\tilde{r}}_{i} = \ltwonorm(\mathtt{Re}(\mathbf{\tilde{f}}_{i})), \\
		\mathbf{\tilde{f}}_{s} = \ltwonorm(\tilde{F}(\mathbf{x}_{s}^{2})),& \quad
		\mathbf{\tilde{r}}_{s} = \ltwonorm(\mathtt{Re}(\mathbf{\tilde{f}}_{s})), \\
		\mathbf{\tilde{r}}^{m} = \lambda^\prime \mathbf{\tilde{r}}_{i} &+ (1 - \lambda^\prime) \mathbf{\tilde{r}}_{s}.
	\end{aligned}
	\label{eq:mix_feat_forward}
\end{equation}

\vspace{-1mm}

Then the contrastive loss can be represented as 
\begin{equation}
	\ell_{mixlrco} = - \log
	\frac { h(\mathbf{r}^m_i, \mathbf{\tilde{r}}^{m})}
	{ h(\mathbf{r}^m_i, \mathbf{\tilde{r}}_i) + h(\mathbf{r}^m_i, \mathbf{\tilde{r}}_s)+ \sum\limits_{\mathbf{\tilde{r}}_{-} \in M} h(\mathbf{r}^m_i, \mathbf{\tilde{r}}_{-})},
	\label{eq:contrastiveloss}
\end{equation}
\noindent where $h$ follows the definition in Eq.(\ref{eq:orig_contrastive_loss}), $M$ still stores re-represented features of target domain low-confidence samples. This loss can be regarded as the log form of a ($|M|+2$)-way softmax-based classifier that tries to classify $\mathbf{r}_i^m$ as $\mathbf{\tilde{r}}_i^m$.

We call the method combing LRCo and Mixup as MixLRCo, and the final loss is presented as 
\begin{equation}
	\begin{aligned}
		\mathcal{L}_{mixlrco} = \mathcal{L}_{strong} + \lambda_{co} \mathbb{E}_{\mathbf{x} \in \mathcal{D}_{s} \cup \mathcal{D}_{tu}^l}  \ell_{mixlrco}.
	\end{aligned}
	\label{eq:mixlrco_all}
\end{equation}
\noindent where $\lambda_{co}$ is kept the same as in Eq. (\ref{eq:lrco_all}).

%% file: tables/uda_officehome.tex
\begin{table*}[ht]
	\setlength{\tabcolsep}{8.0pt} 
	\renewcommand{\arraystretch}{0.95} 
	\caption{\small
	Classification accuracy (\%) of different UDA methods on Office-Home with ResNet-50 as backbone.}
	\vspace{-3mm}
	\label{tab:uda_officehome}
		\resizebox{\textwidth}{!}{
			\begin{tabular}{c@{}|cccccccccccc|c}
				\toprule
				Method& A$\rightarrow$C & A$\rightarrow$P & A$\rightarrow$R & C$\rightarrow$A & C$\rightarrow$P & C$\rightarrow$R & P$\rightarrow$A & P$\rightarrow$C & P$\rightarrow$R & R$\rightarrow$A & R$\rightarrow$C & R$\rightarrow$P & Acc \\
				\hline
				Source-Only  & 34.9 & 50.0 & 58.0 & 37.4 & 41.9 & 46.2 & 38.5 & 31.2 & 60.4 & 53.9 & 41.2 & 59.9 & 46.1 \\
				GVB (CVPR'20)~\cite{cui2020gvb} & 57.0 & 74.7 & 79.8 & 64.6 & 74.1 & 74.6 & 65.2 &  55.1 & 81.0 & 74.6 & 59.7 & 84.3 & 70.4 \\
				HDAN (NeurIPS'20)~\cite{jin2020heuristicDA} & 56.8 & 75.2 & 79.8 & 65.1 & 73.9 & 75.2 & 66.3 & 56.7 & 81.8 & 75.4 & 59.7 & 84.7 & 70.9 \\
				FixBi (CVPR'21)~\cite{na2021fixbi} & 58.1 & 77.3 & 80.4 & 67.7 & \bf{79.5} & 78.1 & 65.8 & 57.9 & 81.7 & 76.4 & 62.9 & \textbf{86.7} & 72.7  \\
				ATDOC (CVPR'21)~\cite{Liang_adtoc} & 60.2 & 77.8 & 82.2 & 68.5 & 78.6 & 77.9 & 68.4 & 58.4 & 83.1 & 74.8 & 61.5 & 87.2 & 73.2 \\
				MetaAlign(CVPR'21)~\cite{wei2021metaalign} & $59.3$ & $76.0$ & $80.2$ & $65.7$ & $74.7$ & $75.1$ & $65.7$ & $56.5$ & $81.6$ & $74.1$ & $ 61.1 $ & $85.2$ & $71.3$ \\
				SCDA (ICCV'21)~\cite{Li_2021_ICCV_SCDA} & 60.7 &  76.4  & 82.8 &  \bf{69.8} &  77.5 & 78.4 &  68.9 & 59.0 & 82.7 & 74.9 & 61.8 & 84.5 & 73.1 \\
				TCM (ICCV'21)~\cite{Yue_2021_ICCV_TCM} & 58.6 & 74.4 & 79.6 & 64.5 & 74.0 & 75.1 & 64.6 & 56.2 & 80.9 & 74.6 & 60.7 & 84.7 & 70.7 \\
			  	ToAlign (NeurIPS'21)~\cite{wei2021toalign} & 57.9 & 76.9 & 80.8 & 66.7 & 75.6 & 77.0 & 67.8 & 57.0 & 82.5 & 75.1 & 60.0 & 84.9 & 72.0 \\ 
			  	CST (NeurIPS'21)~\cite{liu2021cycle_selftraining} & 59.0 & 79.6 & \bf{83.4} & 68.4 & 77.1 & 76.7 & 68.9 & 56.4 & 83.0 & 75.3 & 62.2 & 85.1 & 73.0 \\
				\hline
				\hline
				GVB*  & 57.4 & 72.6 & 79.3 & 63.3 & 73.1 & 74.2 & 64.7 & 54.8 & 81.6 & 74.1 & 61.6 & 84.9 & 70.1 \\
				Strong Baseline & 61.0 & 77.6 & 80.4 & 65.6 & 76.8 & 74.5 & 66.5 & \textbf{60.3} & 83.3 & 76.7 & 64.8 & 85.1 & 72.7 \\
				LRCo (Ours) & 63.3 & 80.8 & 81.5 & 67.9 & 78.5 & \textbf{78.8} & 67.2 & 60.1 & 84.2 & \textbf{77.7} & 66.9 & 85.7 & 74.4 \\	
				\bf{MixLRCo} (Ours) & \textbf{64.4} & \textbf{81.1} & 81.6 & 68.5 & 78.9 & \textbf{78.8} & \textbf{69.1} & 59.9 & \textbf{87.0} & 77.3 & \textbf{67.7} & \textbf{86.7} & \textbf{75.1} \\
				\bottomrule
		\end{tabular}}
		\vspace{-3mm}
\end{table*}

%% file: sec_experiments.tex
\section{Experiments}
\label{sec:experiments}

\subsection{Datasets and Scenarios}
\noindent\textbf{\textit{Unsupervised DA:}} \textit{Office-Home}~\cite{Office-HOME} is a challenging dataset for visual domain adaptation with 15,500 images in 65 categories. It has four significantly different domains: Artistic, Clipart, Product, and Real-World~(abbr. R, C, A, and P). \textit{VisDA-2017}~\cite{VisDA2017} is a large-scale dataset for synthetic-to-real domain adaptation. It contains 152,397 synthetic images for the source domain and 55,388 real-world images for the target domain.

\noindent\textbf{\textit{Semi-Supervised DA:}}  
\textit{DomainNet}~\cite{peng2019moment} is initially a multi-source domain adaptation benchmark. Four domains are involved, \ie, Real, Clipart, Painting, and Sketch~(abbr. R, C, P, and S). Each of them contains images of 126 categories. \textit{Office-Home} consists of Real, Clipart, Art, and Product~(abbr. R, C, A, and P) domains with 65 classes. For both datasets, the number of labeled target data is 1-shot or 3-shot per class.

\subsection{Implementation Details}

For the baseline model, we choose GVB~\cite{cui2020gvb} for UDA task and MME~\cite{saito2019mme} for SSDA task. For the temperature $T_{ce}$, we set it $0.03$ for GVB to achieve comparable performance, and $0.05$ for MME by following the original paper. The hyperparameters (denoted as $\lambda_{align}$ in Eq.(\ref{eq:baseline})) in the baseline method are directly used without any modification. As for the backbone network, we use ResNet50~\cite{ResNet1} for UDA, and AlexNet~\cite{krizhevsky2012alexnet} and ResNet34~\cite{ResNet1} for SSDA.

For the strong baseline, an hyperparameters is probability threshold $\tau$ in Eq.(\ref{eq:fixmatch}). We set $\tau=0.8$ for DomainNet and $\tau=0.9$ for Office-Home in the SSDA task selected by validation set. And for UDA task, we empirically control the proportion of high-confidence samples in 60\%–80\%, where $\tau \in [0.93,0.98]$ is used. For LRCo, there are three hyperparameters: temperature $T_{co}$ in Eq. (\ref{eq:orig_contrastive_loss}), memory bank size $|M|$, and trade-off parameter $\lambda_{co}$ in Eq.(\ref{eq:lrco_all}). We set $T_{co}=0.3$, $|M| = 512$, $\lambda_{co}=0.5$ for all experiments. For MixLRCo, the additional hyperparameter is $\alpha$ of $\mathtt{Beta}$ distribution, and we set $\alpha=1.0$ for both tasks.

\input{./tables/uda_visda_resnet50}

\input{./tables/ssda_officehome_3_shot}

\subsection{Comparison with State-of-the-Art}
In this section, we compare our proposed method with other state-of-the-art methods for UDA and SSDA. Here \textit{Source Only} in UDA task means the model trained only using labeled source data, and $S+T$ in SSDA means the model trained by the labeled source and target data. The \textit{Strong Baseline} represents the model denoted in Eq.~(\ref{eq:strong_baseline}) combining FixMatch-KLD and the used baselines (\ie, GVB~\cite{cui2020gvb} and MME~\cite{saito2019mme}). \textbf{LRCo} represents the model corresponding to Eq.~(\ref{eq:lrco_all}), and \textbf{MixLRCo} represents the model corresponding to Eq.~(\ref{eq:mixlrco_all}).

\vspace{-4mm}

\paragraph{Unsupervised Domain Adaptation.}  Table~\ref{tab:uda_officehome} and Table~\ref{tab:uda_visda} show the results on Office-Home and VisDA-2017, respectively. Although the cosine similarity is adopted, our re-implemented GVB can achieve comparable performance with the original one. 
%
%
There exists inferiority of our method in certain adaptation cases compared with other SOTA methods. It is normal that different methods has advantages in certain adaptation scenarios and drawbacks in others owing to the diversity of adaptation tasks. Since our methods is built on existing baseline, the weaknesses of baseline method are inherited to our LRCo and MixLRCo. However, for the average performance, our LRCo and MixLRCo can achieve SOTA performance.

\vspace{-4mm}

\paragraph{Semi-supervised Domain Adaptation.} Table~\ref{tab:ssda_officehome_3_shot} and Table~\ref{tab:ssda_domainnet} show the results on Office-Home and DomainNet, respectively. The baseline method MME is directly adopted since our architecture is the same as MME.
Compared with SOTA methods, the Strong Baseline achieves comparable performance in Office-Home and relatively lower performance in larger-scale DomainNet. And our LRCo and MixLRCo can achieve new SOTA performance on both tasks.
%

\subsection{Analysis and Discussion}
\paragraph{Sample selection for contrastive Loss.} We particularly consider different samples (\ie, high or low confidence samples) when computing contrastive loss in Sec.~\ref{sec:lrco}.
Table~\ref{tab:ana_high_or_low} gives the results of different selections. In particular, the experiment $0$ represents the Strong Baseline, and the experiment $5$ is our LRCo. When the high-confidence samples appear in both positive and negative pairs (\#1, 3, 7, 9), to better avoid semantic conflict, we discard the high-confidence samples in memory bank that share the same pseudo label as the high-confidence query sample. It can be seen that when adopting high-confidence samples as positive pairs (\#1-3), the performance is always decreased. When the positives pairs are from low-confidence samples, the high-confidence negative samples in memory bank will hurt the performance (\#4, 6). This is mainly because the high-confidence negative samples are closer to the class prototypes and the contrastive loss pushes the low-confidence samples away from them. When the positives pairs are from both low and high confidence samples (\#7, 8, 9), the results cannot exceed our proposed LRCo (\#5).

\input{./tables/ana_high_or_low}

\vspace{-4mm}

\input{./tables/ssda_domainnet}

\paragraph{Necessity of re-representation.} We investigate the necessity of feature re-representation in domain adaptation by comparing it with other choices. Table~\ref{tab:ana_orig_or_rerep} shows the ablation results.%
%
It can be seen that our re-representation mechanism can extremely boost the performance.
We also test re-representation without detaching classifier weights (\ie, update them with contrastive loss), and the performance will drop.

\input{./tables/ana_orig_or_rerep} 

\vspace{-4mm}

\begin{figure}[!t]
	\centering
	\includegraphics[width=0.8\columnwidth]{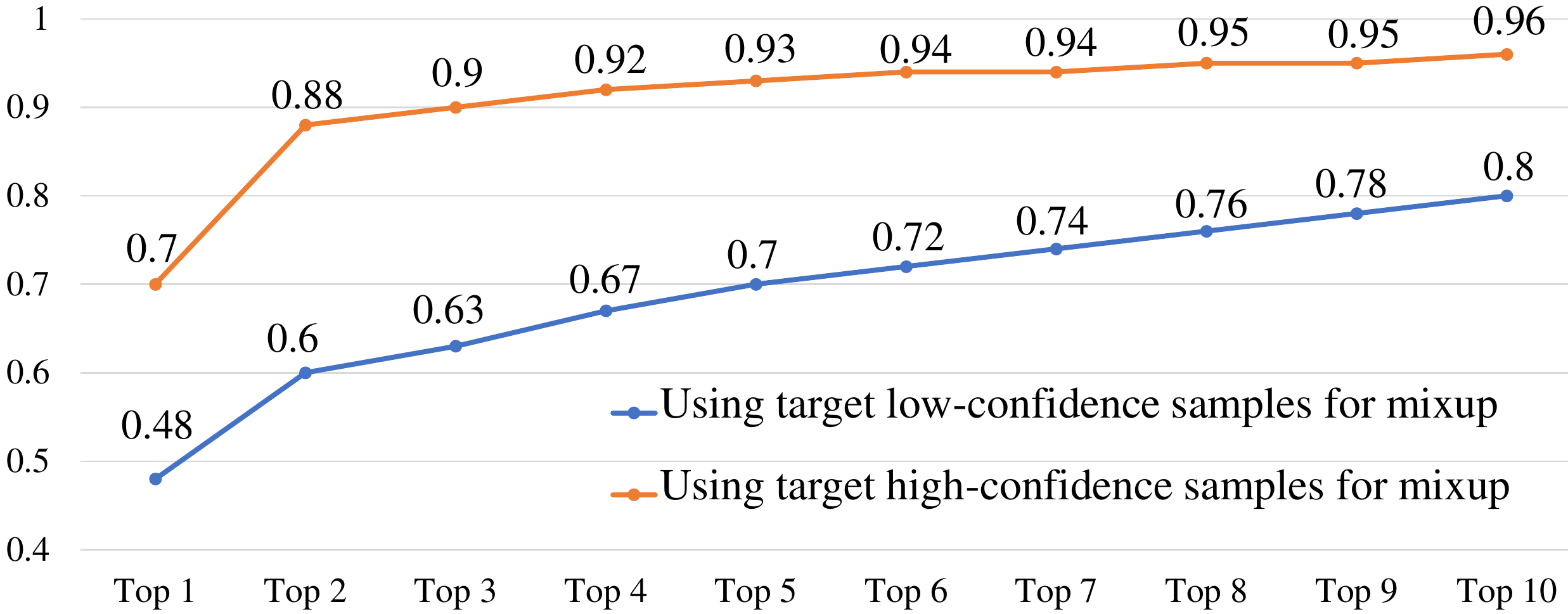}
	\caption{\small The accumulation of largest $k \in \{1,2...10\}$ probabilities of different mixed samples.}
	\label{fig:topk}
\end{figure}

\paragraph{Different designs for MixLRCo.} Our implementation for MixLRCo is non-trial since there are two key designs: \textit{a)} only select the low-confidence samples in target domain for cross-domain mixup. \textit{b)} keep the target domain samples dominant.
The results of different choices are shown in Table~\ref{tab:ana_designs_for_mix}. It can be seen that involving the target high-confidence samples always brings negative effect. When only using the target low-confidence samples, MixLRCo without target domination can achieve better performance than LRCo but lower than MixLRCo with target domination.
One may ask that the mixup of target high-confidence samples and source samples can also produce less confident samples, and why it performs worse. Here we explain it from the probabilities of different mixup samples. Fig.~\ref{fig:topk} shows the average value of the accumulation of top $k \in \{1,2,...10\}$ largest probabilities. It can be seen that the mixup with target high-confidence samples can reach $88\%$ by adding the largest two probabilities, while the mixup with target low-confidence samples only reaches $80\%$ by adding top $10$ largest probabilities. This indicates that these two kinds of mixup samples have distinct distributions, and the mixup samples by high-confidence ones are still confident, which is undesirable for our proposed LRCo as discussed above.

\input{./tables/ana_designs_for_mix}

\begin{figure}[!ht]
	\centering
	\includegraphics[width=0.8\columnwidth]{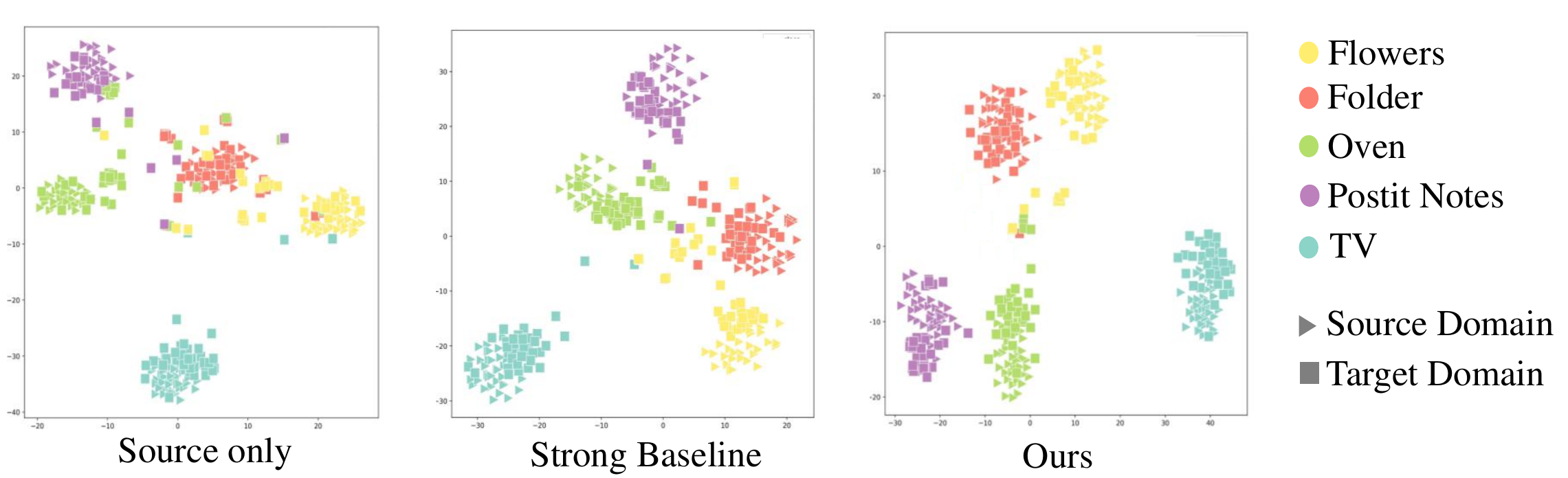}
	\caption{\small The feature TSNE visualization under UDA Office-Home R$\rightarrow$C.}
	\label{fig:tsne}
\end{figure}

\paragraph{Feature Visualization.} As shown in Fig~\ref{fig:tsne}, the target features of our method are more compact and better aligned with source features.

%% file: tables/uda_visda_resnet50.tex
\begin{table}[t]
        \footnotesize
		\centering
		\setlength{\tabcolsep}{20.0pt} 
		\renewcommand{\arraystretch}{0.9} 
		\caption{\small Accuracies (\%) of Synthetic $\rightarrow$ Real on VisDA-2017 for unsupervised domain adaptation methods using ResNet-50.} 
		\begin{tabular}{c|c}
			\toprule
			Method & Acc \\
			\hline
			DANN(ICML'15)~\cite{ganin2015dann}  &57.4\\
			MDD (ICML'19)~\cite{zhang2019MDD}  & 74.6 \\
			MDD + FixMatch~\cite{liu2021cycle_selftraining} & 77.8 \\
			CDAN (NeurIPS'18)~\cite{long2018CDAN}  & 70.0 \\
			CDAN + VAT + Entropy~\cite{liu2021cycle_selftraining} & 76.5 \\
			 GVB (CVPR'20)~\cite{cui2020gvb} & 75.3 \\
			 SENTRY (ICCV'21)~\cite{prabhu2021sentry} & 76.7 \\
			 CST (NeurIPS'21)~\cite{liu2021cycle_selftraining} & 79.9 \\
			 CST + SAM (NeurIPS'21) ~\cite{liu2021cycle_selftraining} & 80.6 \\
			\hline
			\hline
			GVB* & 74.5 \\
			Strong Baseline &  79.3 \\
			LRCo (Ours) &  81.3 \\
			\bf{MixLRCo} (Ours) & \textbf{82.0} \\
			\bottomrule
	\end{tabular}
		\label{tab:uda_visda}
	\vspace{-5mm}
\end{table}

%% file: tables/ssda_officehome_3_shot.tex
\begin{table*}[ht]
\centering
\setlength{\tabcolsep}{10.0pt} 
\renewcommand{\arraystretch}{1.0} 
\vspace{-5mm}
\caption{Accuracy(\%) on \textit{Office-Home} under the setting of 3-shot using Alexnet (A) and Resnet34 (R) as backbone networks.}
\resizebox{\linewidth}{!}{
\begin{tabular}{c|c@{}|cccccccccccc|c}
\toprule
Net & Method & R$\rightarrow$C & R$\rightarrow$P & R$\rightarrow$A & P$\rightarrow$R & P$\rightarrow$C & P$\rightarrow$A & A$\rightarrow$P & A$\rightarrow$C & A$\rightarrow$R & C$\rightarrow$R & C$\rightarrow$A & C$\rightarrow$P & Mean \\
\hline
\multirow{10}{*}{A} & S+T & 44.6 & 66.7 & 47.7 & 57.8 & 44.4 & 36.1 & 57.6 & 38.8 & 57.0 & 54.3 & 37.5 & 57.9 & 50.0 \\
 & MME (ICCV'19)~\cite{saito2019mme} & 51.2 & 73.0 & 50.3 & 61.6 & 47.2 & 40.7 & 63.9 & 43.8 & 61.4 & 59.9 & 44.7 & 64.7 & 55.2 \\
 & APE (ECCV'20)~\cite{Kim2020AttractPA} & 51.9 & 74.6 & 51.2 & 61.6 & 47.9 & 42.1 & 65.5 & 44.5 & 60.9 & 58.1 & 44.3 & 64.8 & 55.6 \\
 & CLDA (NeurIPS'21)~\cite{singh2021clda} & 51.5 & 74.1 & 54.3 & 67.0 & 47.9 & \bf{47.0} & 65.8 & 47.4 & \bf{66.6} & 64.1 & \bf{46.8} & 67.5 & 58.3 \\
 & CDAC (CVPR'21)~\cite{li2021cdac} &  54.9 & 75.8 & 51.8 & 64.3 & 51.3 & 43.6 & 65.1 & 47.5 & 63.1 & 63.0 & 44.9 & 65.6 & 56.8 \\
 & ECACL-P (ICCV'21)~\cite{Li_2021_ICCV_ecacl} & 55.4 & 75.7 & \bf{56.0} & 67.0 & 52.5 & 46.4 & 67.4 & 48.5 & 66.3 & 60.8 & 45.9 & 67.3 & 59.1 \\
 \cline{2-15}
  & MME (ICCV'19)~\cite{saito2019mme} & 51.2 & 73.0 & 50.3 & 61.6 & 47.2 & 40.7 & 63.9 & 43.8 & 61.4 & 59.9 & 44.7 & 64.7 & 55.2 \\
 & Strong Baseline & 56.4 & 74.8 & 51.6 & 67.1 & 53.0 & 42.8 & 66.3 & 48.9 & 64.1 & 62.3 & 43.0 & 67.5
 & 58.2 \\
 & LRCo (Ours) & \bf{58.7} & 76.9 & 52.8 & 68.9 & \bf{54.8} & 43.5 & 68.9 & 50.1 & 66.0 & 63.5 & 44.3 & 69.1 & 59.8 \\
 & \textbf{MixLRCo (Ours)} & 57.4 & \bf{77.7} & 53.0 & \bf{69.2} & 54.0 & 45.5 & \bf{69.3} & \bf{50.5} & 66.1 & \bf{65.0} & 46.0 & \bf{71.0} & \bf{60.4} \\
\hline
\hline
\multirow{8}{*}{R} & S+T & 55.7 & 80.8 & 67.8 & 73.1 & 53.8 & 63.5 & 73.1 & 54.0 & 74.2 & 68.3 & 57.6 & 72.3 & 66.2 \\
 & APE (ECCV'20)~\cite{Kim2020AttractPA} & 66.4 & 86.2 & 73.4 & 82.0 & 65.2 & 66.1 & 81.1 & 63.9 & 80.2 & 76.8 & 66.6 & 79.9 & 74.0 \\
 & CLDA (NeurIPS'21)~\cite{singh2021clda} & 66.0 & 87.6 & 76.7 & 82.2 & 63.9 & 72.4 & 81.4 & 63.4 & 81.3 & 80.3 & 70.5 & 80.9 & 75.5 \\
 & CDAC (CVPR'21)~\cite{li2021cdac} & 67.8 & 85.6 & 72.2 & 81.9 & 67.0 & 67.5 & 80.3 & 65.9 & 80.6 & 80.2 & 67.4 & 81.4 & 74.8 \\
 \cline{2-15}
  & MME (ICCV'19)~\cite{saito2019mme} & 64.6 & 85.5 & 71.3 & 80.1 & 64.6 & 65.5 & 79.0 & 63.6 & 79.7 & 76.6 & 67.2 & 79.3 & 73.1 \\
  & Strong Baseline & 71.0 & 87.6 & 74.8 & 80.6 & 70.1 & 68.9 & 82.4 & 68.1 & 79.9 & 78.6 & 67.7 & 81.5 & 76.0 \\
  & LRCo (Ours) & 72.1 & 88.5 & 76.1 & 82.9 & 67.4 & 72.4 & 83.9 & 69.1 & 82.8 & 82.6 & 71.0 & 84.5 & 77.8 \\
  & \bf{MixLRCo} (Ours) & \bf{73.1} & \bf{89.6} & \bf{77.0} & \bf{84.2} & \bf{71.3} & \bf{73.5} & \bf{84.5} & \bf{70.2} & \bf{83.2} & \bf{83.1} & \bf{72.0} & \bf{86.0} & \bf{79.0} \\
  \bottomrule
\end{tabular}}
\vspace{-2mm}
\label{tab:ssda_officehome_3_shot}
\end{table*}

%% file: tables/ana_high_or_low.tex
\begin{table}[ht]
    \footnotesize
	\centering
	\setlength{\tabcolsep}{7pt} 
    \renewcommand{\arraystretch}{0.95} 
	\caption{\small Accuracy of different sample selections for positive pairs and negative pairs. R$\rightarrow$C in Office-Home is used for both UDA and SSDA.}
	\vspace{-3mm}
	\footnotesize
	\begin{tabular}{c|cc|cc|c|c}
		\toprule
		\multirow{2}{*}{\#} & \multicolumn{2}{c|}{Positive Pairs} & \multicolumn{2}{c|}{Negative Pairs} & \multirow{2}{*}{ UDA} & \multirow{2}{*}{SSDA} \\ \cline{2-5}
		& High             & Low             & High             & Low             &         &    \\ \hline
		0 &                \multicolumn{4}{c|}{Strong Baseline} 					 &    64.8      &   71.0  \\ \hline
		1 &     \checkmark      &                 &      \checkmark     &                 &     65.2    &    71.2 \\ 
		2 &     \checkmark      &                 &                  &      \checkmark    &       65.6   &    71.5 \\ 
		3 &     \checkmark      &                 &      \checkmark     &      \checkmark    &    65.6      &   71.1 \\ \hline
		4 &                  &     \checkmark     &    \checkmark    &         &     65.9     &    71.6 \\ 
		\bf{5} &     &   \textbf{\checkmark}  &       &   \bf{\checkmark}  & \textbf{66.9} & \textbf{72.1}  \\ 
		6 &                  &     \checkmark     &       \checkmark    &       \checkmark   &     66.2     &  71.9  \\ \hline
		7 &     \checkmark      &     \checkmark     &        \checkmark   &                 &    65.8      &   71.5 \\ 
		8 &     \checkmark      &     \checkmark     &                  &      \checkmark    &     66.0     &   71.8 \\ 
		9 &     \checkmark      &     \checkmark     &      \checkmark     &      \checkmark    &      66.1    &  71.6  \\
		\bottomrule
	\end{tabular}
	\label{tab:ana_high_or_low}
	\vspace{-3mm}
\end{table}

%% file: tables/ssda_domainnet.tex
\begin{table*}[ht]
	\setlength{\tabcolsep}{5pt} 
	\renewcommand{\arraystretch}{0.95} 
	\caption{\footnotesize Accuracy(\%) on \textit{DomainNet} under the settings of 1-shot and 3-shot using Alexnet (A) and Resnet34 (R) as backbone networks.}
	\vspace{-6mm}
	\label{tab:ssda_domainnet}
	\begin{center}
		\resizebox{\linewidth}{!}{
			\begin{tabular}{c|c@{}|cccccccccccccc|cc}
				\toprule
				\multirow{2}{*}{Net} & \multirow{2}{*}{Method} & \multicolumn{2}{c}{R$\rightarrow$C} & \multicolumn{2}{c}{R$\rightarrow$P} & \multicolumn{2}{c}{P$\rightarrow$C} & \multicolumn{2}{c}{C$\rightarrow$S} & \multicolumn{2}{c}{S$\rightarrow$P} & \multicolumn{2}{c}{R$\rightarrow$S} & \multicolumn{2}{c|}{P$\rightarrow$R} & \multicolumn{2}{c}{Mean} \\
				& & 1-shot & 3-shot & 1-shot & 3-shot & 1-shot & 3-shot & 1-shot & 3-shot & 1-shot & 3-shot & 1-shot & 3-shot & 1-shot & 3-shot & 1-shot & 3-shot \\ \hline
				\multirow{10}{*}{A} & S+T & 43.3 & 47.1 & 42.4 & 45.0 & 40.1 & 44.9 & 33.6 & 36.4 & 35.7 & 38.4 & 29.1 & 33.3 & 55.8 & 58.7 & 40.0 & 43.4 \\
				& APE (ECCV'20)~\cite{Kim2020AttractPA} & 47.7 & 54.6 & 49.0 & 50.5 & 46.9 & 52.1 & 38.5 & 42.6 & 38.5 & 42.2 & 33.8 & 38.7 & 57.5 & 61.4 & 44.6 & 48.9 \\
				& CLDA (NeurIPS'21)~\cite{singh2021clda} & 56.3 & 59.9& 56.0& 57.2& 50.8& 54.6& 42.5& 47.3& 46.8& 51.4& 38.0 & 42.7 & 64.4 & 67.0& 50.7 & 54.3 \\
				& CDAC (CVPR'21)~\cite{li2021cdac} &  56.9 & 61.4 & 55.9 &  57.5 &  51.6 &  58.9 &  44.8 &  50.7 &  48.1 &  51.7 &  44.1 &  46.7 &  63.8 &  66.8 &  52.1 & 56.2 \\
				& ECACL-T (ICCV'21)~\cite{Li_2021_ICCV_ecacl} & 56.8 & 62.9    & 54.8  & 58.9   & 56.3  &60.5   & 46.6 & 51.0 & \textbf{54.6} & 51.2     & \bf{45.4} & 48.9   & 62.8 & 67.4    & 53.4 & 57.7 \\
				\cline{2-18}
				& MME (ICCV'19)~\cite{saito2019mme} & 48.9 & 55.6 & 48.0 & 49.0 & 46.7 & 51.7 & 36.3 & 39.4 & 39.4 & 43.0 & 33.3 & 37.9 & 56.8 & 60.7 & 44.2 & 48.2 \\
				& Strong Baseline &  56.9 &  62.2 &  54.3 &  56.8 &  52.1 &  61.3 &  44.8 &  49.7 & 46.0 &  50.5 &  43.5 &  48.0 &  60.4 &  63.8 & 51.1 &  56.0   \\  
				& LRCo (Ours) & 59.8 & 64.0  & 57.1 & 59.1 & 55.5 & 62.8 & \bf{46.9} & 51.4 & 48.7 & \bf{53.2} & 44.4 & 50.0 & 64.0 & 66.9 & 53.7 & 58.2 \\
				& \textbf{MixLRCo (Ours)} & \bf{61.2} & \bf{64.4} & \bf{58.7} & \bf{59.8} & \bf{57.9} & \bf{63.8} & 46.2 & \bf{52.4} & 49.4 & \bf{53.2} & 45.0 & \bf{50.7} & \bf{65.5} & \bf{67.6} & \bf{54.8} & \bf{58.8} \\
				\hline
				\hline
				\multirow{13}{*}{R} & S+T & 55.6 & 60.0 & 60.6 & 62.2 & 56.8 & 59.4 & 50.8 & 55.0 & 56.0 & 59.5 & 46.3 & 50.1 & 71.8 & 73.9 & 56.9 & 60.0 \\
				& HDAN (NeurIPS'20)~\cite{jin2020heuristicDA} & 71.7 & 73.9 & 67.1 & 69.1 & 72.8 & 73.0 & 63.7 & 66.3 & 65.7 & 67.5 & 69.2 & 69.5 & 76.6 & 79.7 & 69.5 & 71.3 \\
				& APE (ECCV'20)~\cite{Kim2020AttractPA} & 70.4 & 76.6 & 70.8 & 72.1 & 72.9 & 76.7 & 56.7 & 63.1 & 64.5 & 66.1 & 63.0 & 67.8 & 76.6 & 79.4 & 67.6 & 71.7 \\
				& CDAC (CVPR'21)~\cite{li2021cdac} &  77.4 &  79.6 &  74.2 &  75.1 & 75.5 & 79.3 & 67.6 & 69.9 &  71.0 &  73.4 &  69.2 &  72.5 &  80.4 &  81.9 &  73.6 &  76.0 \\
				& ATDOC (CVPR'21)~\cite{Liang_adtoc} & 64.6 & 65.9 & 70.7 & 72.2 & 80.3 & 80.8 & 74.0 & 75.2 & 70.2 & 71.2 & 65.7 & 67.7 & 68.5 & 69.4 & 70.6 & 71.8 \\
				& CLDA (NeurIPS'21)~\cite{singh2021clda} & 76.1 & 77.7 & 75.1 & 75.7 & 71.0 & 76.4 & 63.7& 69.7 & 70.2 & 73.7 & 67.1 & 71.1 & 80.1 & 82.9 & 71.9 & 75.3 \\
				& ToAlilgn (NeurIPS'21)~\cite{wei2021toalign} & 73.0 & 75.7 & 72.0 & 72.9 & 71.7 & 75.6 & 63.0 & 66.3 & 69.3 & 71.1 & 64.6 & 66.4 & 80.8 & 83.0 & 70.6 & 73.0 \\
				& DECOTA (ICCV'21)~\cite{yang2021decota} & - & 80.4 & - & 75.2 & - & 78.7 & - & 68.6 & - & 72.7 & - & 71.9 & - & 81.5 & - & 75.6 \\
				& ECACL-P (ICCV'21)~\cite{Li_2021_ICCV_ecacl}  & 75.3 & 79.0 & 74.1 & 77.3 & 75.3 & 79.4 & 65.0 & 70.6 & 72.1 & 74.6  & 68.1 & 71.6 & 79.7 & 82.4  & 72.8 & 76.4 \\
				\cline{2-18}
				& MME (ICCV'19)~\cite{saito2019mme} & 70.0 & 72.2 & 67.7 & 69.7 & 69.0 & 71.7 & 56.3 & 61.8 & 64.8 & 66.8 & 61.0 & 61.9 & 76.1 & 78.5 & 66.4 & 68.9 \\
				& Strong Baseline & 76.4 & 79.7 & 74.2 & 75.0 & 73.0 & 79.8 & 66.7 & 69.0 & 70.1 & 71.1 & 70.3 & 71.0 & 77.8 & 81.7 & 72.6 & 75.3 \\
				& LRCo (Ours) & 78.5 & 81.6 & 76.3 & 76.9 & 75.9 & 81.0 & \bf{69.0} & 74.0 & 73.6 & 74.1 & \bf{73.4} & \bf{75.6} & 80.7 & 82.9 & 75.3 & 78.0  \\
				& \bf{MixLRCo} (Ours) & \textbf{78.7} & \bf{81.9} & \bf{76.9} & \bf{77.5} & \bf{78.3} & \bf{81.2} & 68.5 & \bf{74.4} & 7\bf{4.2} & \bf{75.3} & 72.8 & \bf{75.6} & \bf{81.1} & \bf{83.5} & \bf{75.8} & \bf{78.5}  \\
				\bottomrule
		\end{tabular}}
		\vspace{-7mm}
	\end{center}
\end{table*}

%% file: tables/ana_orig_or_rerep.tex
\begin{table}[ht]
\footnotesize
	\centering
	\setlength{\tabcolsep}{12.0pt} 
	\renewcommand{\arraystretch}{0.95} 
	\caption{\small Accuracy comparison of the original features and re-represented features in contrastive learning. R$\rightarrow$C in Office-Home is used for both UDA and SSDA.} 
	\footnotesize
	\vspace{-4mm}
	\begin{tabular}{c|c|c}
		\toprule
		Method & UDA  & SSDA  \\ \hline
		Strong Baseline &   64.8   &   71.0   \\
		Original feature &   65.7	&   71.4	\\  
		Re-representation (LRCo) w/o detach &   66.2	& 	71.6	\\
		Re-representation (LRCo) &  \textbf{66.9} 	& 	\textbf{72.1}	\\
		\bottomrule
	\end{tabular}
	\label{tab:ana_orig_or_rerep}
	\vspace{-3mm}
\end{table}

%% file: tables/ana_designs_for_mix.tex
\begin{table}[t]
\footnotesize
	\centering
	\setlength{\tabcolsep}{8pt} 
	\renewcommand{\arraystretch}{1.0} 
	\vspace{-1mm}
	\caption{\small Accuracy of different designs for MixLRCo. R$\rightarrow$C in Office-Home is used for both UDA and SSDA.}
	\footnotesize
        \begin{tabular}{c|ccc|c|c}
		\toprule
		\multirow{2}{*}{\#} & \multicolumn{2}{c|}{Target Samples}                                             & \multirow{2}{*}{\begin{tabular}[c]{@{}c@{}}Target\\ Domination\end{tabular}} & \multirow{2}{*}{ UDA}  & \multirow{2}{*}{SSDA} \\ \cline{2-3}
		& \multicolumn{1}{c|}{High}    & \multicolumn{1}{c|}{Low}       &                &      &      \\ \hline
		0        & \multicolumn{3}{c}{LRCo}     & 66.9 & 72.1 \\ \hline
		1  & \multicolumn{1}{c}{$\checkmark$} & \multicolumn{1}{c|}{\textbf{}}   &     &  65.8    &   71.3   \\
		2    & \multicolumn{1}{c}{$\checkmark$}  & \multicolumn{1}{c|}{}     &  $\checkmark$  &  65.9    &    71.5  \\ \hline
		3     & \multicolumn{1}{c}{}     & \multicolumn{1}{c|}{$\checkmark$}     &     &  67.4    &   72.5   \\ 
		\bf{4}    & \multicolumn{1}{c}{}       & \multicolumn{1}{c|}{$\checkmark$}    &  $\checkmark$ &   \bf{67.7}   &  \bf{73.1}      \\ \hline
		5    & \multicolumn{1}{c}{$\checkmark$}      & \multicolumn{1}{c|}{$\checkmark$}   &     &    66.7  &  72.1     \\ 
		6    & \multicolumn{1}{c}{$\checkmark$}      & \multicolumn{1}{c|}{$\checkmark$}   &  $\checkmark$ &  67.0    &   72.3   \\
		\bottomrule
		\end{tabular}
	\label{tab:ana_designs_for_mix}
	\vspace{-3mm}
\end{table}

%% file: MixLRCo.bbl
\begin{thebibliography}{10}\itemsep=-1pt

\bibitem{Ben2010A}
Shai Ben-David, John Blitzer, Koby Crammer, Alex Kulesza, Fernando Pereira, and
  Jennifer~Wortman Vaughan.
\newblock A theory of learning from different domains.
\newblock {\em Machine Learning}, 2010.

\bibitem{ben2007analysis}
Shai Ben-David, John Blitzer, Koby Crammer, and Fernando Pereira.
\newblock Analysis of representations for domain adaptation.
\newblock In {\em NeurIPS}, 2007.

\bibitem{berthelot2019mixmatch}
David Berthelot, Nicholas Carlini, Ian Goodfellow, Nicolas Papernot, Avital
  Oliver, and Colin~A Raffel.
\newblock Mixmatch: A holistic approach to semi-supervised learning.
\newblock In {\em NeurIPS}, 2019.

\bibitem{chen2019PFAN}
Chaoqi Chen, Weiping Xie, Wenbing Huang, Yu Rong, Xinghao Ding, Yue Huang,
  Tingyang Xu, and Junzhou Huang.
\newblock Progressive feature alignment for unsupervised domain adaptation.
\newblock In {\em CVPR}, 2019.

\bibitem{chen2017deeplab}
Liang-Chieh Chen, George Papandreou, Iasonas Kokkinos, Kevin Murphy, and Alan~L
  Yuille.
\newblock Deeplab: Semantic image segmentation with deep convolutional nets,
  atrous convolution, and fully connected crfs.
\newblock {\em TPAMI}, 2017.

\bibitem{chen2020simple}
Ting Chen, Simon Kornblith, Mohammad Norouzi, and Geoffrey Hinton.
\newblock A simple framework for contrastive learning of visual
  representations.
\newblock In {\em ICML}, 2020.

\bibitem{chen19closerfewshot}
Wei-Yu Chen, Yen-Cheng Liu, Zsolt Kira, Yu-Chiang Wang, and Jia-Bin Huang.
\newblock A closer look at few-shot classification.
\newblock In {\em ICLR}, 2019.

\bibitem{cicek2019rca}
Safa Cicek and Stefano Soatto.
\newblock Unsupervised domain adaptation via regularized conditional alignment.
\newblock In {\em ICCV}, 2019.

\bibitem{cui2020gvb}
Shuhao Cui, Shuhui Wang, Junbao Zhuo, Chi Su, Qingming Huang, and Qi Tian.
\newblock Gradually vanishing bridge for adversarial domain adaptation.
\newblock In {\em CVPR}, 2020.

\bibitem{deng2019clusterwithteacher}
Zhijie Deng, Yucen Luo, and Jun Zhu.
\newblock Cluster alignment with a teacher for unsupervised domain adaptation.
\newblock In {\em ICCV}, 2019.

\bibitem{dwibedi2021little}
Debidatta Dwibedi, Yusuf Aytar, Jonathan Tompson, Pierre Sermanet, and Andrew
  Zisserman.
\newblock {With a Little Help From My Friends: Nearest-Neighbor Contrastive
  Learning of Visual Representations}.
\newblock In {\em ICCV}, 2021.

\bibitem{french2018selfensemble}
Geoffrey French, Michal Mackiewicz, and Mark Fisher.
\newblock Self-ensembling for visual domain adaptation.
\newblock In {\em ICLR}, 2018.

\bibitem{ganin2015dann}
Yaroslav Ganin and Victor Lempitsky.
\newblock Unsupervised domain adaptation by backpropagation.
\newblock In {\em ICML}, 2015.

\bibitem{gu2020sphericalDA}
Xiang Gu, Jian Sun, and Zongben Xu.
\newblock Spherical space domain adaptation with robust pseudo-label loss.
\newblock In {\em CVPR}, 2020.

\bibitem{he2020momentum}
Kaiming He, Haoqi Fan, Yuxin Wu, Saining Xie, and Ross Girshick.
\newblock Momentum contrast for unsupervised visual representation learning.
\newblock In {\em CVPR}, 2020.

\bibitem{ResNet1}
K. He, X. Zhang, S. Ren, and J. Sun.
\newblock Deep residual learning for image recognition.
\newblock In {\em CVPR}, 2016.

\bibitem{hinton2015distilling}
Geoffrey Hinton, Oriol Vinyals, and Jeff Dean.
\newblock Distilling the knowledge in a neural network.
\newblock {\em Computer ence}, 2015.

\bibitem{hjelm2019learning}
R~Devon Hjelm, Alex Fedorov, Samuel Lavoie-Marchildon, Karan Grewal, Phil
  Bachman, Adam Trischler, and Yoshua Bengio.
\newblock Learning deep representations by mutual information estimation and
  maximization.
\newblock In {\em ICLR}, 2019.

\bibitem{hu2021pronoun}
Dapeng Hu, Jian Liang, Qibin Hou, Hanshu Yan, and Yunpeng Chen.
\newblock Adversarial domain adaptation with prototype-based normalized output
  conditioner.
\newblock {\em IEEE Transactions on Image Processing}, 2021.

\bibitem{2020Bidirectional}
Pin Jiang, Aming Wu, Yahong Han, Yunfeng Shao, and Bingshuai Li.
\newblock Bidirectional adversarial training for semi-supervised domain
  adaptation.
\newblock In {\em IJCAI}, 2020.

\bibitem{jin2020heuristicDA}
Xuan Jin, Shuhui Wang, Yuan He, Qingming Huang, et~al.
\newblock Heuristic domain adaptation.
\newblock In {\em NeurIPS}, 2020.

\bibitem{jin2020MCC}
Ying Jin, Ximei Wang, Mingsheng Long, and Jianmin Wang.
\newblock Minimum class confusion for versatile domain adaptation.
\newblock In {\em ECCV}, 2020.

\bibitem{kalantidis2020hardmix}
Yannis Kalantidis, Mert~Bulent Sariyildiz, Noe Pion, Philippe Weinzaepfel, and
  Diane Larlus.
\newblock Hard negative mixing for contrastive learning.
\newblock In {\em NeurIPS}, 2020.

\bibitem{CAN2019}
Guoliang Kang, Lu Jiang, Yi Yang, and Alexander~G Hauptmann.
\newblock Contrastive adaptation network for unsupervised domain adaptation.
\newblock In {\em CVPR}, 2019.

\bibitem{kim2020cross_domain_ssl}
Donghyun Kim, Kuniaki Saito, Tae-Hyun Oh, Bryan~A Plummer, Stan Sclaroff, and
  Kate Saenko.
\newblock Cross-domain self-supervised learning for domain adaptation with few
  source labels.
\newblock {\em arXiv preprint arXiv:2003.08264}, 2020.

\bibitem{kim2019selftraining}
Seunghyeon Kim, Jaehoon Choi, Taekyung Kim, and Changick Kim.
\newblock Self-training and adversarial background regularization for
  unsupervised domain adaptive one-stage object detection.
\newblock In {\em CVPR}, 2019.

\bibitem{Kim2020AttractPA}
Taekyung Kim and Changick Kim.
\newblock Attract, perturb, and explore: Learning a feature alignment network
  for semi-supervised domain adaptation.
\newblock In {\em ECCV}, 2020.

\bibitem{krizhevsky2012alexnet}
Alex Krizhevsky, Ilya Sutskever, and Geoffrey~E Hinton.
\newblock Imagenet classification with deep convolutional neural networks.
\newblock In {\em NeurIPS}, 2012.

\bibitem{kumar2020gradualDA}
Ananya Kumar, Tengyu Ma, and Percy Liang.
\newblock Understanding self-training for gradual domain adaptation.
\newblock In {\em ICML}, 2020.

\bibitem{lee2021imix}
Kibok Lee, Yian Zhu, Kihyuk Sohn, Chun-Liang Li, Jinwoo Shin, and Honglak Lee.
\newblock {\$}i{\$}-mix: A domain-agnostic strategy for contrastive
  representation learning.
\newblock In {\em ICLR}, 2021.

\bibitem{li2020online}
Da Li and Timothy Hospedales.
\newblock Online meta-learning for multi-source and semi-supervised domain
  adaptation.
\newblock In {\em ECCV}, 2020.

\bibitem{li2021cdac}
Jichang Li, Guanbin Li, Yemin Shi, and Yizhou Yu.
\newblock Cross-domain adaptive clustering for semi-supervised domain
  adaptation.
\newblock In {\em CVPR}, 2021.

\bibitem{Li_2021_ICCV_ecacl}
Kai Li, Chang Liu, Handong Zhao, Yulun Zhang, and Yun Fu.
\newblock Ecacl: A holistic framework for semi-supervised domain adaptation.
\newblock In {\em ICCV}, 2021.

\bibitem{li2021BCDM}
Shuang Li, Fangrui Lv, Binhui Xie, Chi~Harold Liu, Jian Liang, and Chen Qin.
\newblock Bi-classifier determinacy maximization for unsupervised domain
  adaptation.
\newblock In {\em AAAI}, 2021.

\bibitem{Li_2021_ICCV_SCDA}
Shuang Li, Mixue Xie, Fangrui Lv, Chi~Harold Liu, Jian Liang, Chen Qin, and Wei
  Li.
\newblock Semantic concentration for domain adaptation.
\newblock In {\em ICCV}, 2021.

\bibitem{li2019bidirectional}
Yunsheng Li, Lu Yuan, and Nuno Vasconcelos.
\newblock Bidirectional learning for domain adaptation of semantic
  segmentation.
\newblock In {\em CVPR}, 2019.

\bibitem{Liang_adtoc}
Jian Liang, Dapeng Hu, and Jiashi Feng.
\newblock Domain adaptation with auxiliary target domain-oriented classifier.
\newblock In {\em CVPR}, 2021.

\bibitem{liu2019transferabletat}
Hong Liu, Mingsheng Long, Jianmin Wang, and Michael Jordan.
\newblock Transferable adversarial training: A general approach to adapting
  deep classifiers.
\newblock In {\em ICML}, 2019.

\bibitem{liu2021cycle_selftraining}
Hong Liu, Jianmin Wang, and Mingsheng Long.
\newblock Cycle self-training for domain adaptation.
\newblock In {\em NeurIPS}, 2021.

\bibitem{long2018CDAN}
Mingsheng Long, Zhangjie Cao, Jianmin Wang, and Michael~I Jordan.
\newblock Conditional adversarial domain adaptation.
\newblock In {\em NeurIPS}, 2018.

\bibitem{long2017deep}
Mingsheng Long, Han Zhu, Jianmin Wang, and Michael~I Jordan.
\newblock Deep transfer learning with joint adaptation networks.
\newblock In {\em ICML}, 2017.

\bibitem{mao2019virtualmixup}
Xudong Mao, Yun Ma, Zhenguo Yang, Yangbin Chen, and Qing Li.
\newblock Virtual mixup training for unsupervised domain adaptation.
\newblock {\em arXiv preprint arXiv:1905.04215}, 2019.

\bibitem{mei2020iast}
Ke Mei, Chuang Zhu, Jiaqi Zou, and Shanghang Zhang.
\newblock Instance adaptive self-training for unsupervised domain adaptation.
\newblock In {\em ECCV}, 2020.

\bibitem{mishra2021surprisinglyss}
Samarth Mishra, Kate Saenko, and Venkatesh Saligrama.
\newblock Surprisingly simple semi-supervised domain adaptation with
  pretraining and consistency.
\newblock In {\em BMVC}, 2021.

\bibitem{misra2020self}
Ishan Misra and Laurens van~der Maaten.
\newblock Self-supervised learning of pretext-invariant representations.
\newblock In {\em CVPR}, 2020.

\bibitem{miyato2018VAT}
Takeru Miyato, Shin-ichi Maeda, Masanori Koyama, and Shin Ishii.
\newblock Virtual adversarial training: a regularization method for supervised
  and semi-supervised learning.
\newblock {\em TPAMI}, 2018.

\bibitem{na2021fixbi}
Jaemin Na, Heechul Jung, Hyung~Jin Chang, and Wonjun Hwang.
\newblock Fixbi: Bridging domain spaces for unsupervised domain adaptation.
\newblock In {\em CVPR}, 2021.

\bibitem{oord2018representation}
Aaron van~den Oord, Yazhe Li, and Oriol Vinyals.
\newblock {Representation Learning With Contrastive Predictive Coding}.
\newblock {\em arXiv preprint arXiv:1807.03748}, 2018.

\bibitem{pan2020unsupervised_intra_da}
Fei Pan, Inkyu Shin, Francois Rameau, Seokju Lee, and In~So Kweon.
\newblock Unsupervised intra-domain adaptation for semantic segmentation
  through self-supervision.
\newblock In {\em CVPR}, 2020.

\bibitem{pan2019TPN}
Yingwei Pan, Ting Yao, Yehao Li, Yu Wang, Chong-Wah Ngo, and Tao Mei.
\newblock Transferrable prototypical networks for unsupervised domain
  adaptation.
\newblock In {\em CVPR}, 2019.

\bibitem{peng2019moment}
Xingchao Peng, Qinxun Bai, Xide Xia, Zijun Huang, Kate Saenko, and Bo Wang.
\newblock Moment matching for multi-source domain adaptation.
\newblock In {\em ICCV}, 2019.

\bibitem{VisDA2017}
Xingchao Peng, Ben Usman, Neela Kaushik, Judy Hoffman, Dequan Wang, and Kate
  Saenko.
\newblock Visda: The visual domain adaptation challenge.
\newblock In {\em arXiv preprint arXiv:1710.06924}, 2017.

\bibitem{prabhu2021sentry}
Viraj Prabhu, Shivam Khare, Deeksha Kartik, and Judy Hoffman.
\newblock Sentry: Selective entropy optimization via committee consistency for
  unsupervised domain adaptation.
\newblock In {\em ICCV}, 2021.

\bibitem{ren2015faster}
Shaoqing Ren, Kaiming He, Ross Girshick, and Jian Sun.
\newblock Faster r-cnn: Towards real-time object detection with region proposal
  networks.
\newblock In {\em NeurIPS}, 2015.

\bibitem{roy2019feature-whitening}
Subhankar Roy, Aliaksandr Siarohin, Enver Sangineto, Samuel~Rota Bulo, Nicu
  Sebe, and Elisa Ricci.
\newblock Unsupervised domain adaptation using feature-whitening and consensus
  loss.
\newblock In {\em CVPR}, 2019.

\bibitem{sahoo2021contrastmix}
Aadarsh Sahoo, Rutav Shah, Rameswar Panda, Kate Saenko, and Abir Das.
\newblock Contrast and mix: Temporal contrastive video domain adaptation with
  background mixing.
\newblock In {\em NeurIPS}, 2021.

\bibitem{saito2019mme}
Kuniaki Saito, Donghyun Kim, Stan Sclaroff, Trevor Darrell, and Kate Saenko.
\newblock Semi-supervised domain adaptation via minimax entropy.
\newblock In {\em ICCV}, 2019.

\bibitem{saito2017asymmetric_tri}
Kuniaki Saito, Yoshitaka Ushiku, and Tatsuya Harada.
\newblock Asymmetric tri-training for unsupervised domain adaptation.
\newblock In {\em ICML}, 2017.

\bibitem{saito2018maximummcd}
Kuniaki Saito, Kohei Watanabe, Yoshitaka Ushiku, and Tatsuya Harada.
\newblock Maximum classifier discrepancy for unsupervised domain adaptation.
\newblock In {\em CVPR}, 2018.

\bibitem{shen2020unmix}
Zhiqiang Shen, Zechun Liu, Zhuang Liu, Marios Savvides, Trevor Darrell, and
  Eric Xing.
\newblock Un-mix: Rethinking image mixtures for unsupervised visual
  representation learning.
\newblock {\em arXiv preprint arXiv:2003.05438}, 2020.

\bibitem{shu2018dirt}
Rui Shu, Hung Bui, Hirokazu Narui, and Stefano Ermon.
\newblock A dirt-t approach to unsupervised domain adaptation.
\newblock In {\em ICLR}, 2018.

\bibitem{singh2021clda}
Ankit Singh.
\newblock Clda: Contrastive learning for semi-supervised domain adaptation.
\newblock In {\em NeurIPS}, 2021.

\bibitem{fixmatch}
Kihyuk Sohn, David Berthelot, Chun-Liang Li, Zizhao Zhang, Nicholas Carlini,
  Ekin~D Cubuk, Alex Kurakin, Han Zhang, and Colin Raffel.
\newblock Fixmatch: Simplifying semi-supervised learning with consistency and
  confidence.
\newblock In {\em NeurIPS}, 2020.

\bibitem{sun2016return}
Baochen Sun, Jiashi Feng, and Kate Saenko.
\newblock Return of frustratingly easy domain adaptation.
\newblock In {\em AAAI}, 2016.

\bibitem{sun2016deep}
Baochen Sun and Kate Saenko.
\newblock Deep coral: Correlation alignment for deep domain adaptation.
\newblock In {\em ECCV}, 2016.

\bibitem{tang2020srdc}
Hui Tang, Ke Chen, and Kui Jia.
\newblock Unsupervised domain adaptation via structurally regularized deep
  clustering.
\newblock In {\em CVPR}, 2020.

\bibitem{tarvainen2017mean}
Antti Tarvainen and Harri Valpola.
\newblock Mean teachers are better role models: Weight-averaged consistency
  targets improve semi-supervised deep learning results.
\newblock In {\em NeurIPS}, 2017.

\bibitem{tzeng2017adversarialadda}
Eric Tzeng, Judy Hoffman, Kate Saenko, and Trevor Darrell.
\newblock Adversarial discriminative domain adaptation.
\newblock In {\em CVPR}, 2017.

\bibitem{tzeng2014deep}
Eric Tzeng, Judy Hoffman, Ning Zhang, Kate Saenko, and Trevor Darrell.
\newblock Deep domain confusion: Maximizing for domain invariance.
\newblock {\em arXiv preprint arXiv:1412.3474}, 2014.

\bibitem{Office-HOME}
Hemanth Venkateswara, Jose Eusebio, Shayok Chakraborty, , and Sethuraman
  Panchanathan.
\newblock Deep hashing network for unsupervised domain adaptation.
\newblock In {\em CVPR}, 2017.

\bibitem{verma2019manifoldmix}
Vikas Verma, Alex Lamb, Christopher Beckham, Amir Najafi, Ioannis Mitliagkas,
  David Lopez-Paz, and Yoshua Bengio.
\newblock Manifold mixup: Better representations by interpolating hidden
  states.
\newblock In {\em ICML}, 2019.

\bibitem{vu2019advent}
Tuan-Hung Vu, Himalaya Jain, Maxime Bucher, Matthieu Cord, and Patrick
  P{\'e}rez.
\newblock Advent: Adversarial entropy minimization for domain adaptation in
  semantic segmentation.
\newblock In {\em CVPR}, 2019.

\bibitem{wang2019transferableattention}
Ximei Wang, Liang Li, Weirui Ye, Mingsheng Long, and Jianmin Wang.
\newblock Transferable attention for domain adaptation.
\newblock In {\em AAAI}, 2019.

\bibitem{wei2021metaalign}
Guoqiang Wei, Cuiling Lan, Wenjun Zeng, and Zhibo Chen.
\newblock Metaalign: Coordinating domain alignment and classification for
  unsupervised domain adaptation.
\newblock In {\em CVPR}, 2021.

\bibitem{wei2021toalign}
Guoqiang Wei, Cuiling Lan, Wenjun Zeng, Zhizheng Zhang, and Zhibo Chen.
\newblock Toalign: Task-oriented alignment for unsupervised domain adaptation.
\newblock In {\em NeurIPS}, 2021.

\bibitem{wu2020dualmixup}
Yuan Wu, Diana Inkpen, and Ahmed El-Roby.
\newblock Dual mixup regularized learning for adversarial domain adaptation.
\newblock In {\em ECCV}, 2020.

\bibitem{xie2018mstn}
Shaoan Xie, Zibin Zheng, Liang Chen, and Chuan Chen.
\newblock Learning semantic representations for unsupervised domain adaptation.
\newblock In {\em ICML}, 2018.

\bibitem{xu2020gpa}
Minghao Xu, Hang Wang, Bingbing Ni, Qi Tian, and Wenjun Zhang.
\newblock Cross-domain detection via graph-induced prototype alignment.
\newblock In {\em CVPR}, 2020.

\bibitem{xu2020domainmixup}
Minghao Xu, Jian Zhang, Bingbing Ni, Teng Li, Chengjie Wang, Qi Tian, and
  Wenjun Zhang.
\newblock Adversarial domain adaptation with domain mixup.
\newblock In {\em AAAI}, 2020.

\bibitem{xu2019largernorm}
Ruijia Xu, Guanbin Li, Jihan Yang, and Liang Lin.
\newblock Larger norm more transferable: An adaptive feature norm approach for
  unsupervised domain adaptation.
\newblock In {\em ICCV}, 2019.

\bibitem{yan2017mind}
Hongliang Yan, Yukang Ding, Peihua Li, Qilong Wang, Yong Xu, and Wangmeng Zuo.
\newblock Mind the class weight bias: Weighted maximum mean discrepancy for
  unsupervised domain adaptation.
\newblock In {\em CVPR}, 2017.

\bibitem{yan2020improvemixup}
Shen Yan, Huan Song, Nanxiang Li, Lincan Zou, and Liu Ren.
\newblock Improve unsupervised domain adaptation with mixup training.
\newblock {\em arXiv preprint arXiv:2001.00677}, 2020.

\bibitem{yang2021decota}
Luyu Yang, Yan Wang, Mingfei Gao, Abhinav Shrivastava, Kilian~Q Weinberger,
  Wei-Lun Chao, and Ser-Nam Lim.
\newblock Deep co-training with task decomposition for semi-supervised domain
  adaptation.
\newblock In {\em ICCV}, 2021.

\bibitem{yue2021pcs}
Xiangyu Yue, Zangwei Zheng, Shanghang Zhang, Yang Gao, Trevor Darrell, Kurt
  Keutzer, and Alberto~Sangiovanni Vincentelli.
\newblock Prototypical cross-domain self-supervised learning for few-shot
  unsupervised domain adaptation.
\newblock In {\em CVPR}, 2021.

\bibitem{Yue_2021_ICCV_TCM}
Zhongqi Yue, Qianru Sun, Xian-Sheng Hua, and Hanwang Zhang.
\newblock Transporting causal mechanisms for unsupervised domain adaptation.
\newblock In {\em ICCV}, 2021.

\bibitem{yun2019cutmix}
Sangdoo Yun, Dongyoon Han, Seong~Joon Oh, Sanghyuk Chun, Junsuk Choe, and
  Youngjoon Yoo.
\newblock Cutmix: Regularization strategy to train strong classifiers with
  localizable features.
\newblock In {\em ICCV}, 2019.

\bibitem{zhang2018mixup}
Hongyi Zhang, Moustapha Cisse, Yann~N Dauphin, and David Lopez-Paz.
\newblock mixup: Beyond empirical risk minimization.
\newblock In {\em ICLR}, 2018.

\bibitem{zhang2021prototypical}
Pan Zhang, Bo Zhang, Ting Zhang, Dong Chen, Yong Wang, and Fang Wen.
\newblock Prototypical pseudo label denoising and target structure learning for
  domain adaptive semantic segmentation.
\newblock In {\em CVPR}, 2021.

\bibitem{zhang2018iCAN}
Weichen Zhang, Wanli Ouyang, Wen Li, and Dong Xu.
\newblock Collaborative and adversarial network for unsupervised domain
  adaptation.
\newblock In {\em CVPR}, 2018.

\bibitem{zhang2019MDD}
Yuchen Zhang, Tianle Liu, Mingsheng Long, and Michael Jordan.
\newblock Bridging theory and algorithm for domain adaptation.
\newblock In {\em ICML}, 2019.

\bibitem{zhang2019domainsymnet}
Yabin Zhang, Hui Tang, Kui Jia, and Mingkui Tan.
\newblock Domain-symmetric networks for adversarial domain adaptation.
\newblock In {\em CVPR}, 2019.

\bibitem{zhang2021rpa}
Yixin Zhang, Zilei Wang, and Yushi Mao.
\newblock Rpn prototype alignment for domain adaptive object detector.
\newblock In {\em CVPR}, 2021.

\bibitem{zhang2021sslanduda}
Yabin Zhang, Haojian Zhang, Bin Deng, Shuai Li, Kui Jia, and Lei Zhang.
\newblock Semi-supervised models are strong unsupervised domain adaptation
  learners.
\newblock {\em arXiv preprint arXiv:2106.00417}, 2021.

\bibitem{zheng2020coarse_to_fine}
Yangtao Zheng, Di Huang, Songtao Liu, and Yunhong Wang.
\newblock Cross-domain object detection through coarse-to-fine feature
  adaptation.
\newblock In {\em CVPR}, 2020.

\bibitem{zhu2021featuretransformation}
Rui Zhu, Bingchen Zhao, Jingen Liu, Zhenglong Sun, and Chang~Wen Chen.
\newblock Improving contrastive learning by visualizing feature transformation.
\newblock In {\em ICCV}, 2021.

\bibitem{zou2018CBST}
Yang Zou, Zhiding Yu, BVK Kumar, and Jinsong Wang.
\newblock Unsupervised domain adaptation for semantic segmentation via
  class-balanced self-training.
\newblock In {\em ECCV}, 2018.

\bibitem{zou2019crst}
Yang Zou, Zhiding Yu, Xiaofeng Liu, BVK Kumar, and Jinsong Wang.
\newblock Confidence regularized self-training.
\newblock In {\em ICCV}, 2019.

\end{thebibliography}
